\documentclass{article}

\usepackage{microtype}
\usepackage{graphicx}
\usepackage{subfigure}
\usepackage{booktabs} 

\usepackage{hyperref}

\usepackage[accepted]{icml2024}

\usepackage{amsmath}
\usepackage{amssymb}
\usepackage{mathtools}
\usepackage{amsthm}

\usepackage[capitalize,noabbrev]{cleveref}

\usepackage{makecell}
\usepackage{booktabs}  
\usepackage{threeparttable}
\usepackage{multirow}
\usepackage{amsfonts} 
\usepackage{booktabs}
\usepackage{epsfig} 
\usepackage{subfigure}
\usepackage{graphicx}
\usepackage{amsmath}
\usepackage{amssymb, amsthm}
\usepackage{booktabs}
\usepackage{xcolor}

\usepackage{lipsum} 
\usepackage{graphicx}
\usepackage{wrapfig}

\DeclareMathOperator*{\argmin}{arg\,min}

\usepackage{algorithm}
\usepackage{algorithmic}
\renewcommand{\algorithmiccomment}[1]{\hfill $\triangleright$ #1}

\theoremstyle{plain}
\newtheorem{theorem}{Theorem}
\newtheorem{proposition}[theorem]{Proposition}

\theoremstyle{definition}
\newtheorem{definition}[theorem]{Definition}

\theoremstyle{remark}

\usepackage[textsize=tiny]{todonotes}
\usepackage{colortbl}

\icmltitlerunning{Imitation Learning from Purified Demonstrations}

\begin{document}

\twocolumn[
\icmltitle{Imitation Learning from Purified Demonstrations}



\icmlsetsymbol{cores}{*}

\begin{icmlauthorlist}
\icmlauthor{Yunke Wang}{wuhan}
\icmlauthor{Minjing Dong}{hk}
\icmlauthor{Yukun Zhao}{wuhan}
\icmlauthor{Bo Du}{wuhan}
\icmlauthor{Chang Xu}{sydney}
\end{icmlauthorlist}

\icmlaffiliation{wuhan}{School of Computer Science, National Engineering Research Center for Multimedia Software, Institute of Artificial Intelligence and Wuhan Institute of Data Intelligence, Wuhan University, China.}
\icmlaffiliation{hk}{Department of Computer Science, City University of Hong Kong, China.}
\icmlaffiliation{sydney}{School of Computer Science, Faculty of Engineering, The University of Sydney, Australia}

\icmlcorrespondingauthor{Bo Du}{dubo@whu.edu.cn}
\icmlcorrespondingauthor{Chang Xu}{c.xu@sydney.edu.au}

\icmlkeywords{Machine Learning, ICML}

\vskip 0.3in
]



\printAffiliationsAndNotice{}  

\begin{abstract}
Imitation learning has emerged as a promising approach for addressing sequential decision-making problems, with the assumption that expert demonstrations are optimal. However, in real-world scenarios, most demonstrations are often imperfect, leading to challenges in the effectiveness of imitation learning. While existing research has focused on optimizing with imperfect demonstrations, the training typically requires a certain proportion of optimal demonstrations to guarantee performance. To tackle these problems, we propose to purify the potential noises in imperfect demonstrations first, and subsequently conduct imitation learning from these purified demonstrations. Motivated by the success of diffusion model, we introduce a two-step purification via diffusion process. In the first step, we apply a forward diffusion process to smooth potential noises in imperfect demonstrations by introducing additional noise. Subsequently, a reverse generative process is utilized to recover the optimal demonstration from the diffused ones. We provide theoretical evidence supporting our approach, demonstrating that the distance between the purified and optimal demonstration can be bounded. Empirical results on MuJoCo and RoboSuite demonstrate the effectiveness of our method from different aspects.
\end{abstract}

\section{Introduction}
Reinforcement Learning (RL) \cite{sutton2018reinforcement,kaelbling1996reinforcement} has achieved significant success in addressing sequential decision-making problems \cite{silver2016mastering,van2016deep,zha2021douzero}. The core tenet of RL lies in the learning of an optimal policy via rewarding an agent's actions during its interaction with the environment. The strategic formulation of reward is essential to the recovery of the best policy. However, the intricacy of reward engineering often poses significant challenges in real-world tasks, leading to the failure of RL algorithms  occasionally~\cite{amodei2016concrete, dulac2019challenges}. In response to these challenges, an alternative approach to policy learning is Imitation Learning (IL) \cite{abbeel2004apprenticeship, hussein2017imitation,cai2021imitation}, a learning framework that utilizes expert behaviors to guide agent learning. 
One fundamental IL approach is Behavioral Cloning (BC) \cite{torabi2018behavioral, sasaki2021behavioral}, in which the agent observes the action of the expert and learns a policy by directly minimizing the action probability discrepancy via supervised learning. This offline training manner has been proven to suffer from compounding error when the agent executes the policy, leading it to drift to new and dangerous states~\cite{xu2020error,xu2021error}. In contrast, Generative Adversarial Imitation Learning (GAIL)~\cite{ho2016generative, fu2017learning, dadashi2020primal,cai2021seeing,wang2023unlabeled} has revealed that imitation learning can be framed as a problem of matching state-action occupancy measures, resulting in a more accurate policy. Based on the framework of Generative Adversarial Nets (GAN) \cite{goodfellow2014generative}, the discriminator in GAIL is introduced to distinguish demonstrations from expert policy and agent policy, yet the agent policy tries its best to generate behaviors that cheat the judgment of the discriminator.

Current imitation learning approaches have shown promising results under the premise that expert demonstrations exhibit high-quality performance. Nonetheless, acquiring optimal demonstrations can often be costly in practical real-world applications. In many cases, the demonstrations available are not optimal, leading to the problem of imperfect demonstrations in imitation learning. Under such a situation, imitation learning algorithms are prone to failure when expert demonstrations are characterized by noise. Hence, the question of how to effectively learn an optimal policy from imperfect demonstrations becomes central to bridging the application gap of imitation learning from simulated environments to real-world tasks. This issue is critical for enhancing the adaptability and effectiveness of imitation learning methods in various practical scenarios.

Confidence-based methods have been shown to be effective when addressing the issue of imperfect demonstrations in imitation learning. In WGAIL~\cite{wang2021learning} and SAIL~\cite{ijcai2021-0434}, confidence estimation for each demonstration is tied to the discriminator during the adversarial training phase. BCND~\cite{sasaki2021behavioral} proposes that the agent's policy itself can estimate confidence. 
Rather than predicting the confidence directly through either policy or discriminator, CAIL~\cite{zhang2021confidence} considers confidence as a learnable parameter and jointly optimizes it with imitation learning methods. The key of confidence-based methods lies in how to derive proper confidence for each expert demonstration. However, most existing methods integrate confidence estimation into the training of IL~\cite{beliaev2022imitation}, leading to a bi-level optimization problem. This bi-level optimization can become unstable and hard to converge during the training of imitation learning, resulting in a breakdown of the confidence estimation. Furthermore, the aforementioned methods are specifically designed to be compatible with either BC or GAIL. This specificity limits their flexibility and restricts their application with other methods.

Rather than integrating the handling of imperfect demonstrations into IL training, we propose an approach where the purification of imperfect demonstrations is performed first. Subsequently, imitation learning is carried out with purified demonstrations. Based on this idea, we introduce Diffusion Purified Imitation Learning (DP-IL), which utilizes the forward and reverse diffusion processes to recover the optimal demonstrations from imperfect ones. By incorporating the diffusion process into denoising imperfect demonstrations, we provide theoretical analysis on the effectiveness of noise elimination during the forward diffusion process as well as the reduction of the gap between the optimal and purified distribution. The distance between the purified and optimal demonstration can also be bounded. We show that the purified demonstrations can be used in both online and offline imitation learning methods. Experimental results in MuJoCo~\cite{todorov2012mujoco} and RoboSuite~\cite{robosuite2020} demonstrate the effectiveness of our method from different aspects.  

\section{Related Work}
\subsection{Imitation Learning from Imperfect Demonstrations}
When dealing with imperfect demonstrations, confidence-based IL methods estimate the weight for imperfect demonstrations to address their importance to agent learning. 2IWIL~\cite{wu2019imitation} and IC-GAIL~\cite{wu2019imitation} first investigate the effectiveness of weighting schemes in imitation learning with imperfect demonstrations. However, these approaches relied on manually labeled confidence, which is challenging to obtain in practical scenarios. To relax this requirement, subsequent works have proposed alternative methods. For instance, DWBC~\cite{xu2022discriminator} and DICE-based methods~\cite{kim2021demodice,chang2021mitigating,ma2022versatile,yu2023offline, li2023theoretical} leverage a small fraction of known optimal demonstrations to infer the weights for the remaining supplementary demonstrations. CAIL~\cite{zhang2021confidence} utilizes partially ranked trajectories to guide confidence estimation. Recent works also focus on estimating confidence without exposing too much prior information. WGAIL~\cite{wang2021learning} and SAIL~\cite{ijcai2021-0434} connect confidence estimation to the discriminator during training. BCND~\cite{sasaki2021behavioral} use the policy network to indicate confidence. However, these approaches have two primary limitations. First, the weight estimation and imitation learning build up a bi-level optimization problem, which can be challenging to converge. Secondly, most methods rely on having a certain proportion of optimal demonstrations, which may not be feasible in practical settings.

Preference-based methods~\cite{christiano2017deep,ibarz2018reward} have also proven effective for policy learning from imperfect demonstrations. Using human preference can avoid complex reward engineering, thus making policy learning more practical. T-REX \cite{brown2019extrapolating} focuses on extrapolating a reward function based on ranked trajectories. By effectively capturing the rankings, the learned reward function provides valuable feedback to guide the agent's learning process. T-REX only requires precise rankings of trajectories, yet does not set constraints on data quality. As a result, T-REX can achieve satisfactory performance even in scenarios where optimal trajectories are unavailable.
D-REX \cite{brown2020better} introduces relaxation to the ranking constraint of T-REX. Initially, it learns a pre-trained policy through behavioral cloning and subsequently generates ranked trajectories by injecting varying levels of noise into the actions. D-REX uses the same way to learn the reward as T-REX with ranked trajectories. To address potential ranking errors, SSRR \cite{chen2021learning} proposes a novel reward function structure. By mitigating the adverse effects arising from ranking inaccuracies, SSRR enhances the reliability of the learning process.

\subsection{Diffusion Model in Imitation Learning}
Diffusion models have demonstrated significant potential in generative tasks. Some works have explored the use of diffusion models either to directly model the policy network or to serve as the discriminator within Adversarial Imitation Learning (AIL) frameworks. For example, policy network in \cite{chi2023diffusion,pearce2022imitating,reuss2023goal} is defined to be a condition diffusion model that refines a noise to the action based on the given state.
DiffAIL~\cite{wang2024diffail} models the state-action pairs as unconditional diffusion models and uses diffusion loss as part of the discriminator’s learning objective. Despite achieving SOTA performance in some benchmark settings, they can not solve imperfect demonstration issues in imitation learning.
In \cite{wang2023diffusion}, the diffusion model is trained with expert demonstrations and used to enhance the generalization of policy trained by BC. Despite both \cite{wang2023diffusion} and our method trains a diffusion model on expert demonstrations first, the diffusion model is used to achieve different tasks.

\section{Preliminary}
Before diving into our method, we briefly review the definition of the Markov Decision Process (MDP) and Imitation Learning with Distribution Matching.
\subsection{Markov Decision Process (MDP)}
MDP is popular for formulating reinforcement learning (RL) \cite{DBLP:books/wi/Puterman94} and imitation learning (IL) problems. An MDP normally consists six basic elements $M=(\mathcal{S}, \mathcal{A}, \mathcal{P}, \mathcal{R}, \gamma, \mu_0)$, where $\mathcal{S}$ is a set of states, $\mathcal{A}$ is a set of actions, $\mathcal{P}:\mathcal{S}\times\mathcal{A}\times\mathcal{S}\rightarrow [0, 1]$ is the stochastic transition probability from current state $s$ to the next state $s'$, $\mathcal{R}:\mathcal{S}\times\mathcal{A}\rightarrow \mathbb{R}$ is the obtained reward of agent when taking action $a$ in a certain state $s$, $\gamma \in [0,1]$ is the discounted rate and $\mu_0:\mathcal{S}\rightarrow[0, 1]$ denotes the initial state distribution. 
\begin{definition}
For any policy $\pi(a|s):\mathcal{S}\rightarrow\mathcal{A}$, there is an one-to-one correspondence between $\pi$ and its occupancy measure $\rho_\pi:\mathcal{S}\times\mathcal{A}\rightarrow[0, 1]$, which is formulated as
\begin{align}
    \rho_\pi(s,a)=(1-\gamma)\pi(a|s)\sum_{t=0}^{\infty}\gamma^t Pr(s_t=s|\pi),
\end{align}
where $Pr(s_t=s|\pi)$ denotes the probability density of state $s$ at timestep $t$ following policy $\pi$.
\end{definition}

\subsection{Imitation Learning via Distribution Matching}
The field of Imitation Learning (IL) is concerned with optimizing an agent's behavior in a given environment by utilizing expert demonstrations. Given expert demonstrations $\mathcal{D}_e$ sampled from the expert policy $\pi_e$, imitation learning methods aim to let the agent policy $\pi_\theta$ replicate the expert behavior. Distribution Matching (DM) approaches attempt to match the agent's state-action distribution $\rho_{\pi_\theta}$ with that of the expert's $\rho_{\pi_e}$ by minimizing the $f$-divergence~\cite{nowozin2016f, ke2019imitation},
\begin{align} \label{eq:dm}
    {\theta}^{*} & = \arg\min_\theta D_f(\rho_{\pi_e}(s,a), \rho_{\pi_\theta}(s,a))
\end{align}
where $f:\mathbb{R}^{+}\rightarrow\mathbb{R}$ is a convex, lower semi-continuous function and satisfies $f(1)=0$. Different choices of $f$-divergence can recover different imitation learning methods. For example, using KL divergence, Jensen-Shannon divergence can recover Behavior Cloning (BC)~\cite{sasaki2021behavioral}, Generative Adversarial Imitation Learning (GAIL)~\cite{ho2016generative}, respectively.

\section{Methodology}
Imitation learning achieves promising results in benchmark tasks with two non-trivial assumptions: (i) Expert demonstrations are sampled from an optimal policy, and (ii) Expert demonstrations are enough to encompass the expert distribution. While these two assumptions may not always hold in practice, our setting falls into the problem of imitation learning with imperfect demonstrations, where we have access to a small fraction of optimal demonstrations and a large fraction of supplementary sub-optimal demonstrations. To tackle this problem, we propose a two-step method named Diffusion Purified Imitation Learning (DP-IL), in which sub-optimal demonstrations are purified via a combination of forward and reverse diffusion process. Furthermore, we provide theoretical results on the optimal reverse point that could achieve best performance.

\subsection{General Objective}
We begin by formalizing the problem of incorporating supplementary sub-optimal demonstrations into imitation learning. For simplicity, we denote the optimal expert distribution as $\rho_{\pi_o}$ and sub-optimal expert distribution as $\rho_{\pi_s}$.
In this setting, we have access to a limited number of optimal demonstrations $\mathcal{D}_o=\{s_i,a_i\}^{n_o}_{i=1}\sim \rho_{\pi_o}(s,a)$, and a large number of supplementary sub-optimal demonstrations $\mathcal{D}_s=\{s_i,a_i\}^{n_s}_{i=1} \sim \rho_{\pi_s}(s,a)$. Our objective is to leverage both types of demonstrations $\mathcal{D}_e=\mathcal{D}_o \cup \mathcal{D}_s$ to learn a good policy $\pi$ for the agent. Typically, $\mathcal{D}_o$ alone is often insufficient for effective policy learning~\cite{kim2021demodice,xu2022discriminator}, and the involvement of $\mathcal{D}_s$ could significantly hurt the optimization of imitation learning algorithms.

To tackle this problem, a potential solution is to purify supplementary sub-optimal demonstrations $\mathcal{D}_s$ under the guidance of optimal demonstrations $\mathcal{D}_o$. Specifically, assuming that $\rho_{\pi_s}(s,a)$ can be purified through a distribution transformation $\mathcal{F}$, the agent policy $\pi_\theta$ can subsequently learn from the purified demonstrations. To summarize, the objective can be formulated as,
\begin{align} \label{eq:rho_obj}
    \min_\theta& \ D_f(\mathcal{F}^\ast(\rho_{\pi_s}(s,a)), \rho_{\pi_\theta}(s,a)) \\
     \mathrm{s.t.} \ \mathcal{F}^\ast = &\argmin_\mathcal{F} \mathcal{H}(\rho_{\pi_o}(s,a), \mathcal{F}(\rho_{\pi_s}(s,a))),\label{eq:rho_obj2}
\end{align}
where $\mathcal{H}$ denotes the distance measurement between distributions. We assume that sub-optimality within supplementary demonstrations mainly come from the potential perturbations $\delta$ during the collections, which form the sub-optimal expert distribution. Thus, taking $\mathcal{F}^\ast$ as a denoising function can be a potential solution to tackle the gap between optimal and sub-optimal expert distributions.

Motivated by the success of diffusion models in adversarial purification~\cite{nie2022diffusion}, we introduce a Diffusion Purified Imitation Learning (DP-IL) algorithm, which eliminates the potential noises in $\rho_{\pi_s}$ through a two-step diffusion process. In the diffusion model, the first step involves a forward diffusion process, where small timestep noise is gradually added to the sub-optimal demonstrations. This results in diffused demonstrations that eliminate potential perturbation patterns in the sub-optimal expert distribution $\rho_{\pi_s}$ while preserving the original semantic information. In the second step, a reverse-time diffusion process is applied to reconstruct the purified demonstrations from the diffused demonstrations.

\subsection{Purification via Diffusion Process}
We formulate $\mathcal{F}^\ast$ as a two-step purification as in the diffusion model, which includes forward and reverse diffusion processes. For the diffusion model, we adopt widely-used DDPM~\cite{ho2020denoising}. For simplicity, we use $x$ to represent the state-action pair $(s,a)$ in the following subsections since we regard it as a whole during the purification process.

\paragraph{Training Diffusion Model with Optimal Demos.}
Since the transformation $\mathcal{F}$ is a combination of forward and reverse diffusion processes, we train a diffusion model $\epsilon_\phi$ on optimal demonstrations $x_o$ from $\rho_{\pi_o}$ first. To simplify the formulation, we concatenate the state-action pair to construct the latent variable $x$ for the diffusion model $\epsilon_\phi(x_i, i)$. Subsequently, we inject noises $\epsilon$ on $x_o$, where $i$ indicates the number of steps of the Markov procedure in the DDPM, which can be viewed as a variable of the level of noise. The goal of the diffusion model is to reverse the diffusion process (\textit{i.e.}, denoise), yielding the learning objective,
\begin{align}
    \min_{\phi} \mathbb{E}_{x_o,\epsilon,i}[\epsilon - \epsilon_\phi(\sqrt{\Bar{\alpha}_i}\cdot x_o + \sqrt{1-\Bar{\alpha}_t}\epsilon,t)]^2,
\end{align}
where $i$ is sampled from a uniform distribution, $\epsilon$ is the Gaussian noise sampled from $\mathcal{N}(0,I)$, $\alpha_i$ is defined to be $\alpha_i=1-\beta_i$ and $\beta_i$ is the noise schedule which monotonically increase with $i$. $\Bar{\alpha}_i$ is defined to be $\Bar{\alpha}_i=\prod_{s=1}^i \alpha_s$. 

\paragraph{Purifying Sub-optimal Demonstrations.}
After we obtain the trained diffusion model $\epsilon_\phi$, we purify sub-optimal demonstrations $x_{s}$ sample from $\rho_{\pi_s}$ with $\epsilon_\phi$. Supposing the inverse point $i_r \in \{0,..., N-1\}$ is the intermediate step that is used in the diffusion purification process, its forward diffusion can be calculated as follows,
\begin{align}
    \hat{x}_{i_r} = \sqrt{\Bar{\alpha}_{i_r}}\cdot x_{s} + \sqrt{1-\Bar{\alpha}_{i_r}}\epsilon, \quad \epsilon\sim\mathcal{N}(0, I).
    \label{eq:forward_ddpm}
\end{align}
Then, the reverse diffusion process is used to denoise $\hat{x}_{i_r}$ to $\hat{x}_{o}$ from timestep $i_r$ to 0. The reverse diffusion at every step is defined as
\begin{align}
    \hat{x}_{i-1}=\frac{1}{\sqrt{\alpha_i}}\Big(\hat{x}_{i} - \frac{1-\alpha_i}{\sqrt{1-\Bar{\alpha}_i}\epsilon_\phi(\hat{x}_{i}, t)} \Big) + \sqrt{\beta_i} z,
    \label{eq:reverse_ddpm}
\end{align}
where $z\sim\mathcal{N}(0, \textit{I})$. Starting from intermediate step $\hat{x}_{i_r}$, we recover the $\hat{x}_0$ from Eq. \ref{eq:reverse_ddpm}, which is the purified demonstration corresponding to $x_s$. 

\paragraph{Choice of optimal $i_r$}
Intuitively, setting a larger $i_r$ can help to smooth the potential noise within sub-optimal demonstrations in the forward diffusion. When $i_r=N-1$, the purified demonstrations can be seen as samples randomly drawn from the diffusion model $\epsilon_\phi$. However, training such a diffusion model $\epsilon_\phi$ exclusively with limited optimal demonstrations may result in generated demonstrations lacking diversity and coverage, encountering similar issues as imitation learning solely from $\mathcal{D}_o$. A properly chosen $i_r$ achieves good trade-off between denoising (Eq. \ref{eq:forward_ddpm}) and maintaining the structure of sub-optimal demonstrations (Eq. \ref{eq:reverse_ddpm}), thus can help the agent achieve better performance.

Notice that DDPM is a discretization of VP-SDE~\cite{song2020score}, and the step $i$ in DDPM is a discrete value that varies from $0$ to $N-1$. In the theoretical analysis below, we extend DDPM to VP-SDE and scale the timestep to $t\in[0,1]$. Hence, the problem is to find an optimal $t_r^\ast$.

\subsection{Theoretical Analysis}
In the above subsection, we propose to utilize a two-step purification to tackle supplementary sub-optimal demonstrations in imitation learning. While the choice of optimal $t_r^\ast$ remains unknown, we further provide a more theoretical analysis on (i) the impact of $t_r^\ast$ on the purification (Theorem \ref{theo1} and Theorem \ref{theorem:2}), and (ii) how different levels of noises affect the choice of $t_r^\ast$ (Theorem \ref{theo3}).

We first analyze the purification effectiveness with respect to $t$ during the forward diffusion process. Considering the optimal expert distribution $\rho_{\pi_o}$ and the sub-optimal one $\rho_{\pi_s}$, we show that the gap between the diffused distribution $\rho_{\pi_o,t}$ and $\rho_{\pi_s,t}$ becomes smaller with increasing timestep $t$, which implies that the potential perturbations can be smoothed via the gradually added noises during the forward process.
\setcounter{theorem}{0}
\begin{theorem}
\label{theo1}
    Let $\{x_t\}_{t\in \{0,1\}}$ be samples in the forward diffusion process. If we denote $\rho_{\pi_o,t}(x)$ and $\rho_{\pi_s,t}(x)$ as the respective distributions of $x_t$ when $x_{o,0}\sim \rho_{\pi_o,t=0}(x)$ and $x_{s,0}\sim \rho_{\pi_s,t=0}(x)$, we then have,
    \begin{small}
    \begin{align}
        \varsigma \leq -\frac{1}{2}\int \rho_{\pi_o,t}(x)\beta_t||\nabla_{x}\log \rho_{\pi_o,t}(x) - \nabla_{x}\log \rho_{\pi_s,t}(x)||^2_2 \mathrm{d}x,
        \label{eq:theo1}
    \end{align}
    \end{small}
    where $\varsigma=\frac{\partial D_{KL}(\rho_{\pi_o,t}(x)||\rho_{\pi_s,t}(x))}{\partial t}$ denotes the derivative of $t$ to the KL divergence between $\rho_{\pi_o,t}(x)$ and $\rho_{\pi_s,t}(x)$.
\end{theorem}
Theorem \ref{theo1} is derived from DiffPure~\cite{nie2022diffusion}. The difference is that we start from the forward diffusion step in DDPM and extend discrete steps to continuous steps.
The \textit{Fisher} divergence $D_{fisher}(p||q)$ between two data distribution $p$ and $q$ is defined to be $\int p||\nabla\log p-\nabla \log q||^2_2$. Therefore, we have $\varsigma\leq -\frac{1}{2}D_{fisher}(\rho_{\pi_o,t}(x)||\rho_{\pi_s,t}(x))\leq 0$. The equality is only satisfied when $\rho_{\pi_o,t=1}(x)=\rho_{\pi_s,t=1}(x)$. Notice that $\varsigma\leq 0$ indicates that the KL divergence between $\rho_{\pi_o,t}(x)$ and $\rho_{\pi_s,t}(x)$ monotonically decreases towards $t=0$ to $t=1$ during the forward diffusion step. In other words, with a relatively larger $t$ in the forward diffusion process, the divergence between diffused optimal and sub-optimal expert distributions can be minimized. 

The monotonic decrease of $D_{KL}(\rho_{\pi_o,t}(x)||\rho_{\pi_s,t}(x))$ towards $t$ implies that the diffused optimal expert distribution and diffused sub-optimal expert distribution become closer as $t$ increases. Suppose there exists an $\kappa$ such that the difference between the diffused optimal expert distribution and diffused sub-optimal expert distribution becomes negligible when $D_{KL}(\rho_{\pi_o,t}(x)||\rho_{\pi_s,t}(x))\leq\kappa$. We can then find a minimum $t_r \in [0,1]$ to achieve this. Starting from $\rho_{\pi_s,t_r}(x)$, we can stochastically recover $\rho_{\pi_o}(x)$ at $t=0$ through the reverse diffusion process. We further provide theoretical evidence that the distance between the purified demonstration and its optimal version can be bounded via the two-step purification.
\begin{theorem} \label{theorem:2}
Supposing the sub-optimal demonstration $x_s$ perturbed by noise $\delta$ compared to the optimal demonstration $x_o$, and $\hat{x}_o$ is a purified demonstration of $x_s$, the L2 distance between $x_o$ and $\hat{x}_o$ (i.e., $\lVert x_o - \hat{x}_o \lVert$) is bounded within the interval $[\max\{0, \lambda\}, \Lambda]$ with a probability of at least $1-\sigma$, where
\begin{align}\label{eq:lower_bound}
    \lambda(\zeta,t_r,\delta,\sigma)&= \zeta(t_r)\tilde{C}_{gd}-\lVert\delta\rVert - \sqrt{e^{2\zeta(t_r)}-1}{C_{\sigma,d}}, \\
\label{eq:upper_bound}
    \Lambda(\zeta,t_r,\delta,\sigma)&= \zeta(t_r)C_{gd} +\lVert\delta\rVert + \sqrt{e^{2\zeta(t_r)}-1}{C_{\sigma,d}},
\end{align}
and $ \tilde{C}_{gd} \leq 2\lVert \nabla_x \log \rho_t(x) \rVert \leq C_{gd}$, $\zeta(t_r)=\int_0^{t_r}\frac{1}{2}\beta_s\mathrm{d}s$, $C_{\sigma,d}=\sqrt{2d(1+\sqrt{\frac{8\ln(2/\sigma)}{d}})}$, $\sigma \geq 2e^{-d/8}$.
\end{theorem}
As shown in Theorem \ref{theorem:2}, the L2 distance between the optimal demonstration $x_o$ and the purified demonstration $\hat{x}_o$ can be bounded by the terms related to the intermediate timestep $t_r$. 
According to Theorem \ref{theo1}, $t_r$ should be set to a relatively large value to guarantee that the potential perturbations are removed in diffused distributions, however, $t_r$ cannot be arbitrarily large.
\begin{proposition}    \label{theo:pro}
    $\lambda(\zeta,t_r,\delta,\sigma)$ and $\Lambda(\zeta,t_r,\delta,\sigma)$ are monotonically increasing with respect to $t_r$.
\end{proposition}
Proposition \ref{theo:pro} suggests that as $t_r$ increases, the bound of $\lVert x_o - \hat{x}_o \rVert$ increases as well. Thus, a relatively small $t_r$ is required to achieve a smaller lower and upper bound to satisfy the minimum L2 distance. Together with the analysis in Theorem \ref{theo1}, there exists a trade-off on $t_r$ between perturbation purification and recovery performance. An optimal timestep $t_r^\ast$ is expected to meet the requirements that the potential perturbation can be eliminated in diffused distribution while the L2 distance with the related optimal demonstration can also be minimized.
\begin{theorem}
\label{theorem:3}
There exists a inflection point $t_p$ that satisfies $\lambda^{\prime\prime}(t_p)=0$, which makes $\lambda(t_r)$ increases smoothly before $t_p$ while increasing rapidly after $t_p$. The optimal $t_r^\ast$ is around the inflection point $t_p$, or is around solution of the equation
\begin{align}
\begin{small}
    -\frac{(\zeta'(t_r))^2}{\zeta''(t_r)} (e^{2\zeta(t_r)}-1)^{-3/2} + (e^{2\zeta(t_r)}-1)^{-1/2} = \frac{\tilde{C}_{gd}}{C_{\sigma,d}}.
    \label{eq:theo3}
\end{small}
\end{align}
With a probability of $1-\sigma$, there is a positive correlation between $\delta$ and $t_r^\ast$.
\label{theo3}
\end{theorem}
As demonstrated in Theorem \ref{theo3}, the lower bound $\lambda(t_r)$ exhibits a smooth increase initially, implying that selecting a larger $t_r$ before the inflection point $t_p$ is unlikely to significantly impact $\lambda(t_r)$. Combined with the analysis in Theorem \ref{theo1} that shows larger $t_r$ would better smooth the perturbation, we make a conclusion that the optimal $t_r^\ast$ should be in the vicinity of the inflection point $t_p$.

Furthermore, Theorem \ref{theo3} establishes a positive correlation between $\lVert\delta\rVert$ and $t_r^\ast$, suggesting that for more substantial perturbations within sub-optimal demonstrations, the optimal $t_r^\ast$ tends to be relatively larger. Empirical results from the experiments also align with and support this observation.

We further provide sensitive analysis of $\sigma$ in Theorem \ref{app:theo5}. The proofs of all theorems are provided in the appendix.

\begin{table*}[!t]  
\label{tab:offline}
\small
\caption{Performance of DP-BC and compared offline imitation learning methods in four MuJoCo tasks with different sub-optimal demonstrations. `L1', `L2' and `L3' denotes different levels of perturbations within sub-optimal demonstrations.
We also provide the best inverse step $i_r^\ast$ of diffusion purification, and we have $i_r=Nt_r$.
} 
\vskip 0.1in
\centering 
        \begin{tabular}{c|r|rrrrr|rr}
        \toprule  
        Task & Demos  & BC-opt & BC-all & BCND & DWBC & DemoDICE & DP-BC & $i_r^\ast$ \cr 
        \midrule
		\multirow{6}{*}{Ant-v2} & D1-L1& 29±18 & -22±33  & 28±28 & -156±95 & -72±8 &\textbf{261}±54 & 100
            \cr
            &D1-L2 & - & 292±88 &  199±37 & 132±39 & -3±14 & \textbf{803}±114 & 50
            \cr
            &D1-L3 & - & 1500±298 & 1951±311 & 1669±244 &  94±4 & \textbf{2547}±118& 10
            \cr
            &D2-L1 &  - & 1089±237 & 1539±150  & \textbf{1545}±128  &  262±152   & 1402±151 & 30
            \cr
            &D2-L2 &  - & 918±232  & 1514±232  &  1725±214  & 480±139   & \textbf{1982}±111 & 10
            \cr
            &D2-L3 &  - & 1356±137 & 2232±324  & 2484±29   &  2127±260  & \textbf{3414}±40 &  5
            \cr
            \midrule
		\multirow{6}{*}{HalfCheetah-v2} 
            &D1-L1 & -254±103 & 1262±202 & 1267±202 & 618±204  & 22±106   & 
            \textbf{1365}±147 & 10
            \cr
            &D1-L2 & - & 1264±255 & 1862±119 & 1069±264 &  622±159 & \textbf{2440}±274& 10 
            \cr
            &D1-L3 & - & 859±233  & 834±249  & 694±190 &  768±233    & \textbf{4042}±80 & 10
            \cr
            &D2-L1  & - & 1314±194 & 1498±246  & 671±178  &  -64±105  &   \textbf{1530}±39 & 30 \cr
            &D2-L2  & - & 2789±14  & 2241±245  & 2676±20  & 1771±158  &  \textbf{2714}±15 & 30
            \cr
            &D2-L3  & - & 1780±296 & 1990±408   &  4722±37  &  4508±59  & \textbf{4748}±40 & 10
            \cr
            \midrule
		\multirow{6}{*}{Walker2d-v2} 
            &D1-L1 & -4±0 & 21±2 & 102±10 & 1039±164 &  365±39   & \textbf{1697}±219 &  10
            \cr
            &D1-L2 & -  & 1551±87 & 1635±319 & 1617±152 &  1560±291  &    \textbf{1722}±297 & 10
            \cr
            &D1-L3 &  - & 815±277  & 192±37  &  2322±369 &  583±108  &    \textbf{3020}±466 & 5
            \cr
            &D2-L1 & - & 500±245 & 686±301  & 1656±186  &  1749±259   & \textbf{2208}±160 &30 \cr
            &D2-L2 & - & 578±273 & 1018±347  &  2114±21  & 1818±342    & \textbf{3162}±20  & 5\cr
            &D2-L3 & - & 702±244 & 2574±358  & 2637±134   & 1765±528    &  \textbf{3076}±205 & 3\cr
            \midrule
            \multirow{6}{*}{Hopper-v2} 
            &D1-L1 & 320±0 & 743±83 & 590±10  & 761±12  &  995±119  & \textbf{1000}±42 & 10
            \cr
            &D1-L2 & - & 2044±157  & 2119±206  & 826±22  & 2059±4 & \textbf{3145}±11  & 5
            \cr
            &D1-L3 & - & 3090±37 & 2883±26 &  1140±27 & \textbf{3207}±7 & 1825±137 & 1
            \cr
            &D2-L1 & - & 2191±118 & 2165±181  & 761±122  & 2336±3   & \textbf{2408}±113  & 10 \cr
            &D2-L2 & - & 2266±3  & 2256±2  &  826±2  &  2265±8   & \textbf{2323}±2 & 10  \cr
            &D2-L3 & - & 2712±6   & 3093±10  & 1140±27   &  \textbf{3306}±2   & 2265±171 & 5  \cr
	\bottomrule
	\end{tabular}
\label{t1} 
\end{table*}
\section{Experiments}
In this section, we conduct extensive experiments to verify the effectiveness of DP-IL in MuJoCo \cite{todorov2012mujoco} and Robosuite~\cite{robosuite2020} with different compared methods. The experimental results demonstrate the advantage of DP-IL from different aspects. 

\noindent\textbf{Benchmarking} \quad
We first conduct experiments on MuJoCo benchmarks in OpenAI Gym \cite{1606.01540}. Four popular tasks  (\textit{i.e.}, Ant-v2, HalfCheetah-v2, Walker2d-v2 and Hopper-v2) are used to evaluate DP-IL. The evaluated performance in MuJoCo benchmark can be measured by the average cumulative ground-truth rewards (\textit{i.e.}, the higher the better). We repeat experiments for 5 trials with different random seeds for common practice. Additionally, to verify the robustness of DP-IL with real-world human operation demonstrations, we also conduct experiments on a robot control task in Robosuite platform.

\noindent\textbf{Source of Demonstrations}\quad
In our setting, we have access to limited number of optimal demonstrations $\mathcal{D}_o$ and a lot supplementary sub-optimal demonstrations $\mathcal{D}_s$. To form $\mathcal{D}_o$, an optimal policy $\pi_o(a|s)$ trained by TRPO~\cite{schulman2015trust} to sample optimal demonstrations. As for $\mathcal{D}_s$, we use two different kinds of ways to generate sub-optimal behaviors following existing works~\cite{tangkaratt2019vild,wu2019imitation}:
\begin{itemize}
\item D1: We add Gaussian noise to optimal action distribution $a^\ast$ of $\pi_o$ to form $\pi_s$. The action of $\pi_s$ is modeled as $a\sim\mathcal{N}(a^\ast,\delta)$ and we choose $\delta=[0.6, 0.4, 0.25]$ to form $\mathcal{D}_s$ with varying quality.
\item D2: We save checkpoints during the RL training as the sub-optimal policy $\pi_s$.
\end{itemize}
In addition, since we use optimal demonstrations $\mathcal{D}_o$ to train the diffusion model, we include $\mathcal{D}_o$ into the training of compared methods for a fair comparison.

\noindent\textbf{Compared Methods}\quad
Purified demonstrations are applied to both the offline imitation learning method (\textit{e.g.}, BC) and the online imitation learning method (\textit{e.g.}, GAIL), to verify the generalization of DP-IL. In the case of offline imitation learning, we compare our method against several state-of-the-art offline methods, including BCND~\cite{sasaki2021behavioral}, DWBC~\cite{xu2022discriminator} and DemoDICE~\cite{kim2021demodice}. For online imitation learning, we compare our method with several confidence-based methods, including 2IWIL/IC-GAIL~\cite{wu2019imitation} and WGAIL~\cite{wang2021learning}. 
Further details (\textit{e.g.}, implementation of DP-IL and compared methods, data quality, and more results) can be found in the appendix.

\subsection{Evaluations on MuJoCo}
We evaluated the effectiveness of DP-BC under the offline IL setting, and results are presented in Table \ref{t1}. As have introduced in the experimental setup, we use two different ways to generate sub-optimal demonstrations (\textit{i.e.}, D1 and D2).
BC-opt means we only use $\mathcal{D}_o$ to conduct BC training, while BC-all means we use both $\mathcal{D}_o$ and $\mathcal{D}_s$ to train the policy network. Our findings align with those results reported in ~\cite{kim2021demodice}, showing that using only $\mathcal{D}_o$ can sometimes yield worse performance compared to BC-all. This can be attributed to the insufficient coverage of the entire optimal demonstration space by a limited number of optimal demonstrations. BC-all benefits from a larger training dataset and can outperform BC-opt in certain cases. However, the challenge of sub-optimal demonstrations still hampers BC's ability to achieve optimal performance. BCND outperforms the baseline at some cases due to its self-weighting strategy. While this strategy is unstable when dealing with a high ratio of sub-optimal demonstrations, we can observe that it sometimes collapses. DWBC and DemoDICE demonstrate great performance with D2 demonstrations with relatively smaller perturbations (\textit{e.g.}, L3), however, we find it is more likely to fail when dealing with sub-optimal demonstrations with strong perturbations.
In contrast, our method purifies sub-optimal demonstrations before conducting BC, which consistently improves the subsequent performance of behavior cloning across all different types of sub-optimal demonstrations compared to other methods. This demonstrates the efficacy of our approach in enhancing the performance of BC. Furthermore, the optimal reverse point $i_r^\ast$ is provided in the table, and the results of DP-BC is related to this $i_r^\ast$. As `L1', `L2' and `L3' represent sub-optimal demonstrations with progressively less perturbation, Theorem \ref{theo3} predicts a corresponding decrease in the optimal reverse point, $i_r^\ast$. This pattern is generally observed, illustrating the theorem's validity. 
The impact of optimal $i_r^\ast$ will be discussed in the following ablation study in Sec \ref{sec:ablation}.
\begin{figure}[!t]
    \centering
          \includegraphics[width=0.40\textwidth]{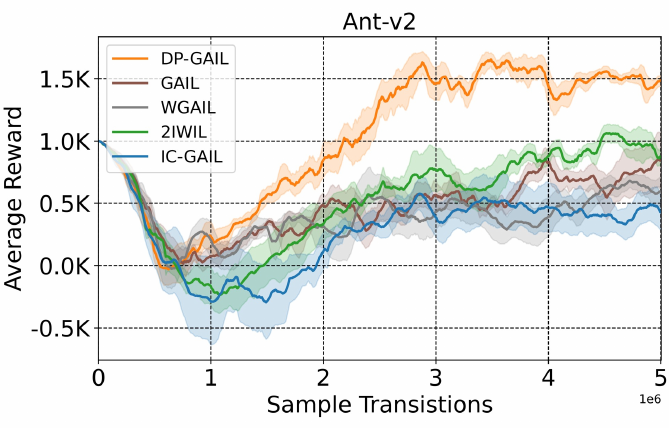}
    \caption{The training curve of DP-GAIL and other online imitation learning methods with D1-L1 demonstrations. The x-axis is the number of interactions with the environment and the shaded area indicates the standard error.}
    \label{fig:fig_gail}
    \vskip -0.1in
\end{figure}

Besides offline imitation learning, we also utilize diffusion-purified demonstrations for GAIL training and compare DP-GAIL with other state-of-the-art online imitation learning methods that also address the imperfect demonstration issue. The training curves of DP-GAIL and other compared methods are depicted in Figure \ref{fig:fig_gail}. The curve unequivocally demonstrates the superiority of DP-GAIL over other compared methods. This shows the efficacy of noise purification in improving the performance of online imitation learning methods. 

\subsection{Ablation Study}
\label{sec:ablation}
\noindent\textbf{Different Types of Noises}\quad
In D1, sub-optimal demonstrations are formed by adding Gaussian noise into the optimal action. It would be interesting to further explore more noises. We consider adding uniform noise and salt-and-pepper noise. Similar to D1, we set $\delta=[0.25,0.4,0.6]$ to form L1 to L3 demonstrations. From the Table \ref{tab:noise}, we can observe that DP-BC performs best in most cases. This suggests the robustness of DP-BC when dealing with different kinds of noises. Intuitively, the noises added to the current data are overwhelmed by the accumulating Gaussian noise during the diffusion forward process, ultimately making the noise less prominent.
\begin{table}[t]  
	\centering 
    \caption{The performance of DP-BC with demonstrations in different noises.
	} 
 \vskip 0.08in
 \small
  \setlength{\tabcolsep}{4.0pt}
\begin{tabular}{l|l|c|c|c}
            \hline\rowcolor{gray!25}
            Task & Demons & BC & BCND & DP-BC\\
            \hline
            \multirow{2}{*}{Ant}
            & Uni-L1
            & 209$\pm$138 & 66$\pm$17 & \bf 343$\pm$16 \\ 
            & Uni-L2
            & 396$\pm$42 & 333$\pm$60 & \bf 480$\pm$15 \\
            & Uni-L3
            & 704$\pm$97 & 1472$\pm$229  & \bf 1925$\pm$227 \\
            \hline
            \multirow{2}{*}{HalfCheetah}
            & Uni-L1
            & 704$\pm$71 & 692$\pm$37 & \bf 733$\pm$8 \\ 
            & Uni-L2
            & 922$\pm$207 & 705$\pm$179  & \bf 1248$\pm$54 \\
            & Uni-L3
            & 1805$\pm$15 & 1933$\pm$21  & \bf  2268$\pm$11 \\
            \hline
            \hline
            \multirow{2}{*}{Ant}
            & S\&P-L1
            & 209$\pm$138 & 686$\pm$45 & \bf 744$\pm$17 \\ 
            & S\&P-L2
            & 396$\pm$42 & 477$\pm$62  & \bf 508$\pm$12 \\
            & S\&P-L3
            & 704$\pm$97 & \bf 755$\pm$143  & 689$\pm$37 \\
            \hline
            \multirow{2}{*}{HalfCheetah}
            & S\&P-L1
            & 2143$\pm$10 & 2042$\pm$113 & \bf 2731$\pm$18 \\ 
            & S\&P-L2
            & 1964$\pm$195 & 1873$\pm$270 & \bf 3550$\pm$18\\
            & S\&P-L3
            & 579$\pm$280 & 1160$\pm$295  & \bf  2820$\pm$383 \\
            \hline
			\end{tabular}
   \vskip -0.07in
\label{tab:noise} 
\end{table}
\begin{table}[t]
    \centering
    \small
    \caption{The performance of DP-BC on Ant-v2 task when defining different $\mathcal{F}^\ast$ in Eq. \ref{eq:rho_obj}. `G-Filter', `M-Filter' and `Med-Filter' denote using Gaussian, Mean and Median Filter as $\mathcal{F}^\ast$ to denoise sub-optimal demonstrations.}
    \vskip 0.05in
    \setlength{\tabcolsep}{5pt} 
    \begin{tabular}{c|cc|ccc}
    \hline\rowcolor{gray!25}
        Demons & BC & DP-BC & G-Filter & M-Filter & Med-Filter \\ \hline
        D1-L1 & -22  & \bf261  & 242  &  182 & 138\\ 
        D1-L2 & 292  & \bf803  & 729  & 736  & 437\\ 
        D1-L3 & 1500  & \bf2547  & 771  & 1786  & 455\\
        \hline
    \end{tabular}
    \vskip -0.1in
    \label{tab:smooth}
\end{table}
\begin{table}[t]
    \centering
    \small
    \caption{The performance of DP-BC with mixed demonstrations (\textit{i.e.}, D1-L1, D1-L2, and D1-L3).}
    \vskip 0.05in
    \setlength{\tabcolsep}{3.8pt} 
    \begin{tabular}{l|cccc|c}
    \hline\rowcolor{gray!25}
        Task & BC & BCND & DWBC & DemoDICE & DP-BC \\ \hline
        Ant & 193  & 411  &  339  & -53  & 601\\ 
        HalfCheetah & 1255   & 921  & 265  & 344  & 3146\\ 
        Walker2d &  169 &  441 & 290  & 488  & 1390\\
        Hopper & 587  & 564  &  541  & 605   & 655 \\
        \hline
    \end{tabular}
    \vskip -0.1in
    \label{tab:mixed}
\end{table}
\begin{figure*}[!t]
    \centering
    \includegraphics[width=1\textwidth]{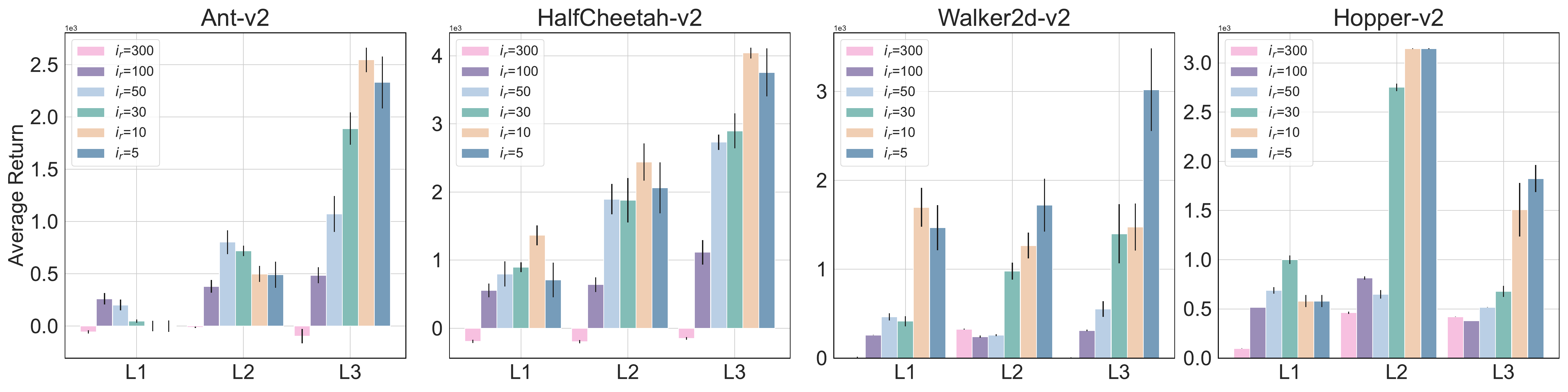}
    \vskip -0.05in
    \caption{Impact of diffusion time $t_r$ with demonstrations of different optimality.
    }
    \vskip -0.1in
    \label{fig:optimal_t}
\end{figure*}

\noindent\textbf{Compared with Other Purified Methods}\quad
The transformation $\mathcal{F(\cdot)}$ is defined as a combination of the forward and reverse diffusion processes to denoise sub-optimal demonstrations. Apart from the diffusion model, there are various denoising methods that can be employed as the transformation $\mathcal{F}^\ast$. Since filters are commonly used to smooth noisy data, we utilize three different filters (mean, median, and Gaussian) as $\mathcal{F}^\ast$ to denoise the imperfect demonstrations. The results are presented in Table \ref{tab:smooth}. It is evident from the table that our method outperforms other filter-based methods in all types of sub-optimal demonstrations (\textit{e.g.}, D1-L1, D1-L2 and D1-L3). Among the filter-based methods, the 'Mean Filter' achieves the best performance. Another notable finding is that when the noise level is high in sub-optimal demonstrations (\textit{e.g.}, L1), filter-based methods tend to maintain a greater advantage over the baseline.

\noindent\textbf{Multiple Demonstrators}\quad
In addition to collecting sub-optimal demonstrations from a single demonstrator, we also investigate the performance of DP-BC when confronted with sub-optimal demonstrations that are sampled from multiple demonstrators. The results are presented in Table \ref{tab:mixed}, where the mixed demonstrations denote a combination of L1, L2, and L3 demonstrations. 
As shown in the table, DP-BC with demonstrations sampled from a mixture of dataset also consistently outperforms the baseline and other compared methods by a substantial margin. 

\noindent\textbf{Impact of optimal $i_r$}\quad
As discussed in Section 4.3, $i_r$ is an important hyper-parameter to achieve the trade-off between smoothing the noise and keeping semantic information. To investigate how the choice of $i_r$ affects the effectiveness of noise purification, we conducted experiments on demonstrations with different levels of optimality. 
Our empirical results confirmed the theoretical findings of Theorem \ref{theo1}, suggesting that demonstrations with less optimality require a relatively larger timestep $t_r$ to smooth the noise. For example, in Ant-v2 task, the optimal $t_r$ is 100 for D1-L1 demonstrations while the optimal $t_r$ is 10 for D1-L3 demonstrations. 
As the quality of demonstrations improved (e.g., from L1 to L3), we observed a gradual decrease in the optimal value of $i_r$. 
This could lead to a smaller value bound since  $\zeta(i_r)$ exhibits a monotonic increase with respect to $t_r$. Hence, diffusion purified imitation learning should perform better when using a smaller $i_r$ under such a case, which is consistent with the results in Figure \ref{fig:optimal_t}.

\subsection{Evaluations on RoboSuite Platform}
We also evaluate the robustness of DP-BC on the RoboSuite platform~\cite{robosuite2020} with real-world demonstrations. We consider a reaching task within the ``Nut Assembly'' in Saywer. In this task, the goal for the robot is to move a nut close to the peg. It earns a higher reward if the nut is closer to the peg. During the reaching phase, the Sawyer robot's arm operates with the gripper in a fully open state, not actively attempting to grasp any objects. Its role is merely to move the nut across the plane. As such, the grasp command is not engaged and prevents the robot's arm from inserting the nut into the peg. Hence, an episode in the reaching task completes only when the timestep matches the horizon length.

\begin{table}[!t]
    \centering
    \small
    \caption{Successful rate of DP-BC in RoboSuite platform with human demonstrations.}
    \vskip 0.1in
    \setlength{\tabcolsep}{6pt} 
    \begin{tabular}{c|c|c|c|c}
    \hline    \rowcolor{gray!25}
    Method    &BC & BCND & DWBC & DP-BC (Ours)  \\
    \hline
    Success Rate& 0.76 & 0.56 &  0.82 & \textbf{0.86} \\
    \hline\hline    \rowcolor{gray!25}
    \bf $i_r$ &1 &3 & 5 & 10  \\
    \hline
    DP-BC & 0.78 & \textbf{0.86} & 0.74 & 0.60 \\
    \hline
    \end{tabular}
    \label{tab:robosuite}
\end{table}
\begin{figure}[t]
    \centering
    \vskip -0.07in
    \includegraphics[width=3in]{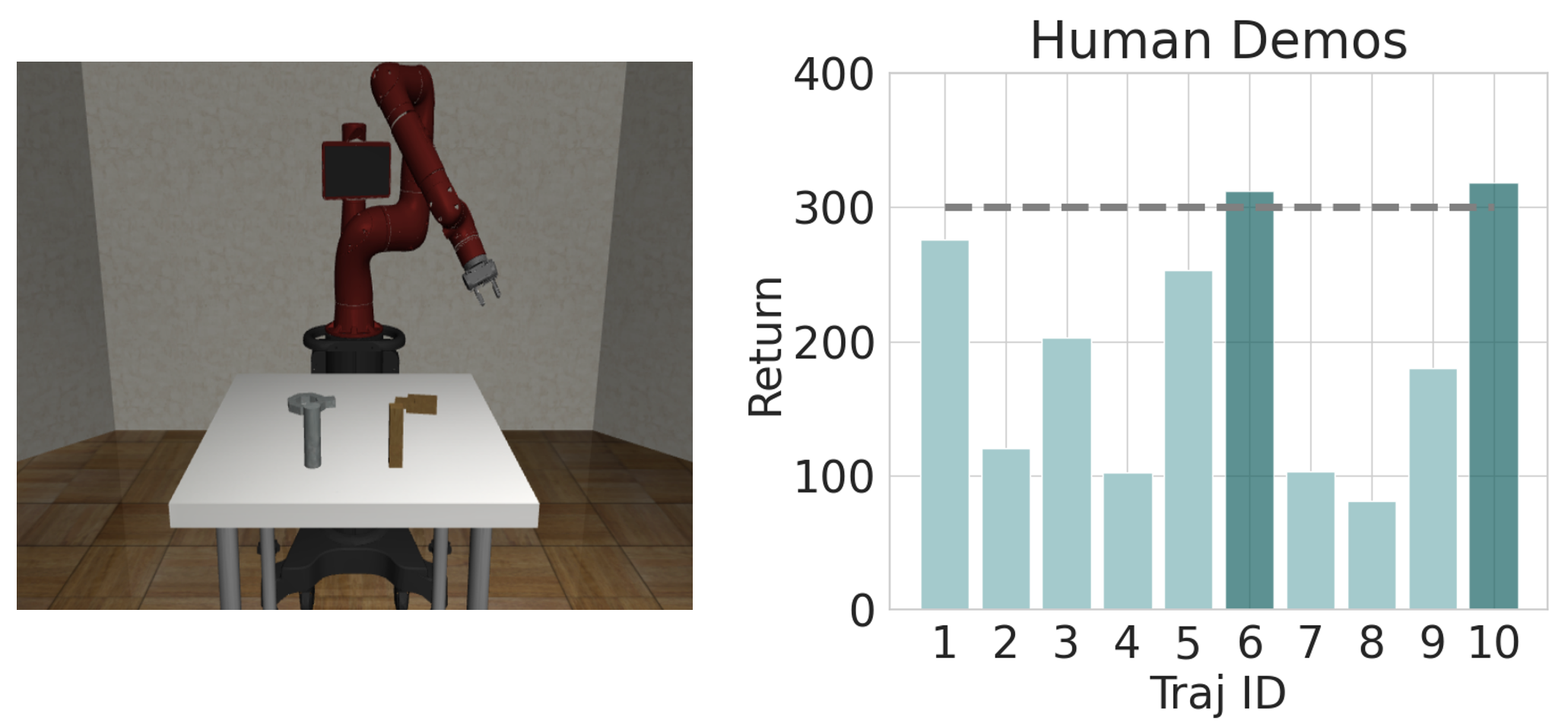}
    \vskip -0.1in
    \caption{Visualization of SaywerNutAssembly task in RoboSuite platform and the quality of human demonstrations.}
    \vskip -0.15in
    \label{fig:robo}
\end{figure}

We use real-world demonstrations by human operators from RoboTurk~\cite{mandlekar2018roboturk}. The demonstrations contain 10 trajectories with approaching length (around 500 state-action pairs). Based on the accumulative reward of trajectories in Figure \ref{fig:robo}, we regard two trajectories that have the larger reward (\textit{e.g.}, $>$300) as optimal demonstrations and the remaining trajectories are regarded as supplementary sub-optimal demonstrations. In this reaching task, we report sucess rate rather than reward to assess learned policy. We set the termination condition of the episode to be when the distance between the nut and the peg is less than 0.1. Based on this criterion, we can provide the success rates of different methods, which are calculated over 50 sampled trajectories. The results are shown in Table \ref{tab:robosuite}, and we can observe that the proposed method has higher success rate than compared methods like BCND and DWBC. We also conduct an ablation study on the timestep $i_r$ in this task and we find that setting $i_r=3$ leads to the best performance.

The results presented in Table \ref{apptab:robosuite} also support our theorem about there exists a trade-off in finding an optimal $i_r$. To further evaluate the relationship between $i_r^\ast$ and different levels of sub-optimal demonstrations, we divided the sub-optimal demonstrations into two groups and evaluate their performance in the appendix (Table \ref{apptab:robosuite}).

\section{Conclusion}
In this paper, we propose to tackle the imperfect demonstrations issue by conducting a two-step purification process to eliminate the potential noises based on the diffusion process. The distance gap between optimal and sub-optimal expert distributions can be minimized after forward diffusion. The purified demonstrations can then be recovered from diffused one via reverse diffusion. Additionally, we provide sufficient theoretical analysis to indicate the impact of the reverse point on the purification. The proposed method can be easily adopted in existing imitation learning frameworks, such as GAIL and BC, to alleviate the effect of sub-optimal expert demonstrations. 
We conduct extensive experiments on MuJoCo and RoboSuite with different types of sub-optimal demonstrations to evaluate the effectiveness of diffusion purification. The comparison results demonstrate the superiority of DP-IL over other baselines.

\section*{Impact Statement}
Our work enables agents to learn from a broader range of imperfect data. There are many potential societal consequences of our work, and the safety of learning from such purified data remains an area worthy of further investigation.

\section*{Acknowledgements}
This work was supported by the National Key Research and Development Program of China 2023YFC2705700, National Natural Science Foundation of China under Grants 62225113, the Innovative Research Group Project of Hubei Province under Grants  2024AFA017, the Australian Research Council under Projects DP210101859 and FT230100549, and the CityU APRC Project No. 9610680.

\bibliography{example_paper}
\bibliographystyle{icml2024}

\clearpage
\onecolumn
\appendix
\setcounter{equation}{11}
\setcounter{theorem}{0}
\section{Proof}
\subsection{Proof of Theorem 1}
\begin{theorem}
\label{app:theo1}
    Let $\{x_t\}_{t\in \{0,1\}}$ be the forward diffusion process in Eq. \ref{eq:forward_ddpm}. If we denote $\rho_{\pi_o,t}(x)$ and $\rho_{\pi_s,t}(x)$ as the respective distributions of $x_t$ when $x_0\sim \rho_{\pi_o,t=0}(x)$ and $x_0\sim \rho_{\pi_s,t=0}(x)$, we then have,
    \begin{small}
    \begin{align}
        \varsigma \leq -\frac{1}{2}\int \rho_{\pi_o,t}(x)\beta_t||\nabla_{x}\log \rho_{\pi_o,t}(x) - \nabla_{x}\log \rho_{\pi_s,t}(x)||^2_2 \mathrm{d}x \leq 0.
    \end{align}
    \end{small}
    where $\varsigma=\frac{\partial D_{KL}(\rho_{\pi_o,t}(x)||\rho_{\pi_s,t}(x))}{\partial t}$ denotes the derivative of $t$ to the KL divergence between $\rho_{\pi_o,t}(x)$ and $\rho_{\pi_s,t}(x)$.
\end{theorem}
\begin{proof}
Following the proof in ~\cite{song2021maximum}, we first make two assumptions on $\rho_{\pi_o,t}(x)$ and $\rho_{\pi_s,t}(x)$. Supposing both $\rho_{\pi_o,t}(x)$ and $\rho_{\pi_s,t}(x)$ are smooth and fast decaying functions, we have
\begin{align}
    \label{app:eq13}
    \lim_{x\rightarrow \infty} \rho_{\pi_o,t}(x) \frac{\partial}{\partial x}\log \rho_{\pi_o,t}(x) = 0 \\
    \lim_{x\rightarrow \infty} \rho_{\pi_s,t}(x) \frac{\partial}{\partial x}\log \rho_{\pi_s,t}(x) = 0
    \label{app:eq14}
\end{align}
The forward diffusion process in Eq. \ref{eq:forward_ddpm} is a discrete Markov chain, which can be written as
\begin{align}
    x_i = \sqrt{1-\beta_i}\cdot x_{i-1} + \sqrt{\beta_i}\cdot \epsilon_i, \quad i=1, ..., N,
    \label{appeq:ddpm_forward}
\end{align}
where $\epsilon_i \sim \mathcal{N}(0, I)$. To obtain a process of continuous transformation in time, we can rewrite Eq. \ref{appeq:ddpm_forward} as
\begin{align}
    x_{(i+1)/N}=\sqrt{1-\frac{\Bar{\beta}_{\frac{i+1}{N}}}{N}}x_{i/N}+\sqrt{\frac{\Bar{\beta}_{\frac{i+1}{N}}}{N}}\epsilon_{(i+1)/N}, \quad i=0, ..., N-1,
    \label{appeq:ddpm_forward1}
\end{align}
\\where   $\{\Bar{\beta}_{i/N}=N{\beta}_i\}^{N-1}_{i=0}$. In the limit of $N\longrightarrow \infty$, $\{\Bar{\beta}_{i/N}=N{\beta}_i\}^{N-1}_{i=0}$ and $\{X_{i/N}\}^{N-1}_{i=0}$ become sequences of function $\{\beta_t\}^{1}_{t=0}$ and $\{x_t\}^{1}_{t=0}$. Let $\Delta t = \frac{1}{N}, t=\frac{i}{N}$, we can rewrite Eq. \ref{appeq:ddpm_forward1} as
\begin{align}
    x_{t+\triangle t}&=\sqrt{1-\beta_{t+\triangle t}\cdot\triangle t}\cdot x_t+\sqrt{\beta_{t+\triangle t}\cdot\triangle t}\cdot\epsilon_t\\
    &=(1 - \frac{1}{2}\beta_{t+\triangle t}\cdot\triangle t)\cdot x_t+\sqrt{\beta_{t+\triangle t}\triangle t}\cdot\epsilon_t + o (\beta_t \cdot \triangle t)
\end{align}
In the limit of $\triangle t \rightarrow 0$, we have 
\begin{align}
    x_{t+\Delta t} - x_t = -\frac{1}{2}\beta_{t}\Delta t x_t + \sqrt{\beta_{t}\Delta t}\cdot \epsilon_t
\end{align}
In other words, we can transform Eq. \ref{eq:forward_ddpm} to the following VP-SDE,
\begin{align}
    \mathrm{d}x = - \frac{1}{2}\beta_t x \mathrm{d}t + \sqrt{\beta_t}\mathrm{d}w,
\end{align}
where $w$ is a standard Wiener process. The Fokker–Planck equation for the VP-SDE above describes the time-evolution of the stochastic process’s associated probability density function, and is given by
\begin{align}
    \frac{\partial \rho_{\pi_o,t}(x)}{\partial t} & =
    - \nabla_{x} \cdot \Big(\frac{1}{2}\beta_t\rho_{\pi_o,t}(x)\nabla_{x}\log\rho_{\pi_o,t}(x) +\frac{1}{2}\beta_t x\rho_{\pi_o,t}(x)
    \Big)\\&= - \nabla_{x}\cdot\Big(h_o(x,t)\rho_{\pi_o,t}(x)
    \Big), 
\end{align}
where $h_o(x,t)$ is defined as $h_o(x,t)=\frac{1}{2}\beta_t\nabla_{x}\log\rho_{\pi_o,t}(x)+\frac{1}{2}\beta_t x$. Then, we denote $\varsigma=\frac{\partial D_{KL}(\rho_{\pi_o,t}(x)||\rho_{\pi_s,t}(x))}{\partial t}$ and $\varsigma$ can be re-written as follows,
\begin{align}
    \varsigma = &
    \frac{\partial}{\partial t}\int \rho_{\pi_o,t}(x)\log \frac{\rho_{\pi_o,t}(x)}{\rho_{\pi_s,t}(x)}\mathrm{d}x \\
    =&
    \int \frac{\partial \rho_{\pi_o,t}(x)}{\partial t}\log \frac{\rho_{\pi_o,t}(x)}{\rho_{\pi_s,t}(x)}\mathrm{d}x  + 
    \int \frac{\partial \rho_{\pi_o,t}(x)}{\partial t}\mathrm{d}x - \int \frac{\rho_{\pi_o,t}(x)}{\rho_{\pi_s,t}(x)}\frac{\partial \rho_{\pi_s,t}(x)}{\partial t}\mathrm{d}x \\
    \overset{(i)}{=}&
    \int \nabla_{x}\cdot\Big(-h_o(x,t)\rho_{\pi_o,t}(x)\Big)\log \frac{\rho_{\pi_o,t}(x)}{\rho_{\pi_s,t}(x)}\mathrm{d}x - \int \frac{\rho_{\pi_o,t}(x)}{\rho_{\pi_s,t}(x)} \nabla_{x}\cdot\Big(-h_s(x,t)\rho_{\pi_s,t}(x)\Big) \\
    \overset{(ii)}{=}&
    - \int \rho_{\pi_o,t}(x)[h_o^T(x,t)-h_s^T(x,t)][\nabla_{x}\log \rho_{\pi_o,t}(x)-\nabla_{x}\log \rho_{\pi_s,t}(x)] \\
    =&
    -\frac{1}{2}\int\rho_{\pi_o,t}(x)\beta_t|\nabla_{x}\log \rho_{\pi_o,t}(x) - \nabla_{x}\log \rho_{\pi_s,t}(x)|^2_2 \mathrm{d}x \\
    =&
    -\frac{\beta_t}{2}E_{x \sim \rho_{\pi_o,t}(x)}(|\nabla_{x}\log \rho_{\pi_o,t}(x) - \nabla_{x}\log \rho_{\pi_s,t}(x)|^2_2) \\
    \leq& \quad 0
\end{align}
where $(i)$ can be obtained by Eq. \ref{app:eq13} and Eq. \ref{app:eq14}, $(ii)$ can be obtained by integration by parts.
\end{proof}

\subsection{Proof of Theorem 2}
\begin{theorem} \label{app:theorem:2}
Supposing the sub-optimal demonstration $x_s$ perturbed by noise $\delta$ compared to the optimal demonstration $x_o$, and $\hat{x}_o$ is a purified demonstration of $x_s$, the L2 distance between $x_o$ and $\hat{x}_o$ (i.e., $\lVert x_o - \hat{x}_o \lVert$) is bounded within the interval $[\max\{0,\lambda\}, \Lambda]$ with a probability of at least $1-\sigma$, where
\begin{align}\label{appeq:lower_bound}
    \lambda(\zeta,t_r,\delta,\sigma)=  \zeta(t_r)\tilde{C}_{gd}-\lVert\delta\rVert - \sqrt{e^{2\zeta(t_r)}-1}{C_{\sigma,d}}, \\
\label{appeq:upper_bound}
    \Lambda(\zeta,t_r,\delta,\sigma)= \zeta(t_r)C_{gd} +\lVert\delta\rVert + \sqrt{e^{2\zeta(t_r)}-1}{C_{\sigma,d}},
\end{align}
and $ \tilde{C}_{gd} \leq 2\lVert \nabla_x \log \rho_t(x) \rVert \leq C_{gd}$, $\zeta(t_r)=\int_0^{t_r}\frac{1}{2}\beta_s\mathrm{d}s$, $C_{\sigma,d}=\sqrt{2d(1+\sqrt{\frac{8\ln\frac{2}{\sigma}}{d}})}$, $\sigma \geq 2e^{-d/8}$.
\end{theorem}
\begin{proof}
Following the proof in \cite{nie2022diffusion}, we extend the distance bound to both upper and lower bounds. The sub-optimal demonstration $x_s$ can be conceptualized as a perturbation of the optimal demonstration $x_o$, with a small disturbance $\delta$ applied to each state-action pair. 
Furthermore, let $x_{t}$ denote the demonstration obtained after subjecting $x_s$ to a forward diffusion process satisfies
\begin{align}
    x_{t} = \sqrt{\gamma({t})}x_s + \sqrt{1-\gamma({t})}\epsilon_1,
\end{align}
where $\gamma(t)=\exp(-\int_0^t\beta_s\mathrm{d}s)$ and $\epsilon_1 \in \mathcal{N}(0,I_d)$, the L2 distance between the purified demonstrations $\hat{x}_o$ and its related optimal demonstration $x_o$ can be upper bounded as
\begin{align}
    \lVert \hat{x}_o-x_o\rVert &= \lVert x_{t}+(\hat{x}_o-x_{t})-x_o\rVert \\
                  &= \lVert x_{t}+ \int_{t}^0 -\frac{1}{2}\beta_t[x+2\nabla_x \log \rho_t(x)]\mathrm{d}t + \int_{t}^0 \sqrt{\beta_t}\mathrm{d}\overline{w} -x_o\rVert \label{theo2:eq2} \\
                  &\leq \lVert x_{t}+ \int_{t}^0 -\frac{1}{2}\beta_t x\mathrm{d}t + \int_{t}^0 \sqrt{\beta_t}\mathrm{d}\overline{w} -x\rVert + \lVert \int_{t}^0 -\beta_t \nabla_x\log \rho_t(x) \mathrm{d}t\rVert,
                  \label{theo2:eq3}
\end{align}
where Eq. \ref{theo2:eq2} follows with the definition of reverse-time diffusion and Eq. \ref{theo2:eq3} can be obtained via triangle inequality.
Similarly, it can be lower bounded as
\begin{align}
    \lVert \hat{x}_o-x_o\rVert 
    \geq&
    \lVert \int_{t}^0 \beta_t\nabla_x\log \rho_t(x)\mathrm{d}t\rVert - \lVert x_{t}+ \int_{t}^0 -\frac{1}{2}\beta_t x\mathrm{d}t + \int_{t}^0 \sqrt{\beta_t}\mathrm{d}\overline{w} -x\rVert  
\end{align}
The sum of the first 3 terms in Eq. \ref{theo2:eq2} is a time-varying Ornstein-Uhlenbeck process with a negative time increment that starts from $t=t$ to $t=0$ with the initial value set to $x_{t}$. 
Denote by $x^\prime_0$ its solution, from ~\cite{sarkka2019applied} we know $x^\prime_0$ follows a Gaussian distribution, where its mean $\mu(0)$ and covariance matrix $\Sigma(0)$ are the solutions of the following two differential equations, respectively,
\begin{align}
    \frac{\mathrm{d}\mu}{\mathrm{d}t} = -\frac{1}{2}\beta_t\mu,  \quad \frac{\mathrm{d}\Sigma}{\mathrm{d}t}=-\beta_t\Sigma + \beta_t I_d,
\end{align}
with the initial conditions $\mu(t)=x_{t}$ and $\Sigma(t)=0$. By solving these two differential equations, we have that conditioned on $x_{t}$, $x^\prime_0\sim \mathcal{N}(e^{\zeta(t_r)}x_{t}, (e^{2\zeta(t_r)}-1)I_d)$, where $\zeta(t_r)=\int_0^{t}\frac{1}{2}\beta_s\mathrm{d}s$. 

Note that,
\begin{align}
    x_{t} =& \sqrt{\gamma(t)}x_s + \sqrt{1-\gamma(t)}\epsilon_1 \\ 
    =& e^{-\zeta(t_r)}(x_o+\delta) + \sqrt{1-e^{-2\zeta(t_r)}}\epsilon_1
\end{align}

Using the reparameterization trick, we have,
\begin{align}
    x^\prime_0-x_o &= e^{\zeta(t_r)}x_{t} + \sqrt{e^{2\zeta(t_r)}-1}\epsilon_2 -x_o \\
                  &= e^{\zeta(t_r)} \Big(e^{-\zeta(t_r)}(x_o+\delta) + \sqrt{1-e^{-2\zeta(t_r)}}\epsilon_1\Big) + \sqrt{e^{2\zeta(t_r)}-1}\epsilon_2 -x_o \\
                  &= \sqrt{e^{2\zeta(t_r)}-1}(\epsilon_1 + \epsilon_2) + \delta \\
                  &= \sqrt{2(e^{2\zeta(t_r)}-1})\epsilon + \delta 
\end{align}
Since $\epsilon_1$ and $\epsilon_2$ are independent and taken from the distribution $\mathcal{N}(0,I_d)$, $\epsilon = \frac{\epsilon_1 + \epsilon_2}{\sqrt{2}} \sim \mathcal{N}(0,I_d)$.
Assuming that $ \tilde{C}_{gd} \leq 2\lVert \nabla_x\log \rho_t(x) \rVert \leq C_{gd}$, we have,
\begin{align}
    \lVert \hat{x}_o-x_o\rVert \leq& \lVert x'_0 - x_o \rVert + \lVert \int_{t_r}^0 -\beta_t\nabla\log \rho_t(x))\mathrm{d}t\rVert \\
    =& \Big\lVert\sqrt{2(e^{2\zeta(t_r)}-1)}\epsilon + \delta \Big\rVert + \zeta(t_r)C_{gd} \\
    \leq& \Big\lVert\sqrt{2(e^{2\zeta(t_r)}-1)}\epsilon \Big\rVert + \lVert \delta \rVert + \zeta(t_r)C_{gd}
\end{align}
Similarly, we also have,
\begin{align}
    \lVert \hat{x}_o-x\rVert
    \geq& \lVert \int_{t_r}^0 -\beta_t\nabla\log \rho_t(x)\mathrm{d}t\rVert - \lVert x'_0 - x \rVert\\
    \geq& 
    \zeta(t_r)\tilde{C}_{gd} -
    \Big\lVert\sqrt{2(e^{2\zeta(t_r)}-1)}\epsilon + \delta \Big\rVert \\
    \geq&
    \zeta(t_r)\tilde{C}_{gd} -
    \Big\lVert\sqrt{2(e^{2\zeta(t_r)}-1)}\epsilon \Big\rVert - \lVert \delta \rVert 
\end{align}
Since $\lVert \epsilon \rVert^2 = \mathbf{Z}^2_1 + \mathbf{Z}^2_2 + \dots + \mathbf{Z}^2_d \sim \mathcal{X}^2(d)$, where $\mathbf{Z}^2_i \sim \mathcal{N}(0,1)$. It can be regarded as a sub-exponential random variable with parameters $(\nu,\alpha) = (2\sqrt{d},4)$, we have
\begin{align}
    Pr \Big(|\frac{1}{d} \sum\limits_{n=1}^{d} \mathbf{Z}_n^2 - 1 | \geq t \Big) \leq 2e^{-dt^2/8}, \quad \text{for all} \quad t \in (0,1)
\end{align}
Let $\sigma = 2e^{-dt^2/8}$, we have
\begin{align}
Pr\Bigg(\sqrt{d(1-\sqrt{\frac{8ln(2/\sigma)}{d}})} \leq \lVert \epsilon \rVert \leq \sqrt{d(1+\sqrt{\frac{8ln(2/\sigma)}{d}})}\Bigg) \geq 1-\sigma
\end{align}
By integrating the results above, we can derive Eqs. \ref{appeq:lower_bound}, \ref{appeq:upper_bound}. These two equations present the lower and upper bounds of the L2 distance between $\hat{x}_o$ and $x_o$ with the probability of at least $1-\sigma$, where $\sigma \geq 2e^{-d/8}$.    
\end{proof}

\subsection{Proof of Proposition 3}
\begin{proposition}    \label{apptheo:pro}
    $\lambda(\zeta,t_r,\delta,\sigma)$ and $\Lambda(\zeta,t_r,\delta,\sigma)$ are monotonically increasing with respect to $t_r$.
\end{proposition}
\begin{proof}
The derivative of $\lambda(t_r)$ with respect to $t_r$ can be written as
\begin{align}
    \lambda'(t_r) = -\zeta'(t_r)(e^{2\zeta(t_r)}-1)^{-1/2}C_{\sigma,d} +\zeta'(t_r)\tilde{C}_{gd}
    \label{appeq:lambda_derivate}
\end{align}
With the assumption that the lower bound is non-negative, we have
\begin{align}
    \tilde{C}_{gd} \geq \frac{1}{\zeta(t_r)} (\lVert \delta \rVert + \sqrt{e^{2\zeta(t_r)}-1}C_{\sigma,d})
        \label{appeq:lambda_derivate1}
\end{align}
By substituting Eq. \ref{appeq:lambda_derivate1} into Eq. \ref{appeq:lambda_derivate}, we have,
\begin{align}
    \lambda'(t_r) \geq \frac{\zeta'(t_r)}{\zeta(t_r)} (e^{2\zeta(t_r)}-1)^{-1/2} [C_{\sigma,d}(e^{2\zeta(t_r)}-\zeta(t_r)-1) + \sqrt{e^{2\zeta(t_r)}-1}\lVert \delta \rVert]
\end{align}
Given that \(\zeta(t_r) \geq 0\), \(\zeta'(t_r) = \beta_{t_r} \geq 0\) and \(e^{2\zeta(t_r)} - \zeta(t_r) - 1 \geq 0\), it follows that \(\lambda'(t_r) \geq 0\). Considering the upper bound \(\Lambda(t_r) = \zeta(t_r)C_{gd} + \|\delta\| + \sqrt{e^{2\zeta(t_r)}-1}C_{\sigma,d}\), and noting that \(\Lambda(t_r)\) increases monotonically as a function of \(\zeta(t_r)\), and since \(\zeta(t_r)\) itself increases monotonically with \(t_r\), it can be deduced that \(\Lambda(t_r)\) is also a monotonically increasing function of \(t_r\).
\end{proof}

\subsection{Proof of Theorem 4}
\begin{theorem}
\label{apptheorem:3}
There exists a inflection point $t_p$ that satisfies $\lambda^{\prime\prime}(t_p)=0$, which makes $\lambda(t_r)$ increases smoothly before $t_p$ while increasing rapidly after $t_p$. The optimal $t_r^\ast$ is around the inflection point $t_p$, or is around the solution of the equation
\begin{align}
\begin{small}
    -\frac{(\zeta'(t_r))^2}{\zeta''(t_r)} (e^{2\zeta(t_r)}-1)^{-3/2} + (e^{2\zeta(t_r)}-1)^{-1/2} = \frac{\tilde{C}_{gd}}{C_{\sigma,d}}.
    \label{appeq:theo3}
\end{small}
\end{align}
With a probability of $1-\sigma$, there is a positive correlation between $\lVert\delta\rVert$ and $t_r^\ast$.
\label{apptheo3}
\end{theorem}
\begin{proof}
We aim to identify a critical point $t_p$ where the rate of growth of the function experiences a significant increase. In mathematical words, $\lambda(t_r)$ should satisfies $\lambda'(t_r) \geq 0$, $\lambda''(t_p)=0$ and $\lambda''(t_r) < 0 \ \text{for} \ t_r<t_p$, $\lambda''(t_r) > 0$ for $t_r>t_p$. The second derivative of $\lambda(t_r)$ can be expressed as:
\begin{align}
    \lambda''(t_r) = \left[-\zeta''(t_r) + (\zeta'(t_r))^2(e^{2\zeta(t_r)}-1)\right](e^{2\zeta(t_r)}-1)^{-1/2}C_{\sigma,d} + \zeta''(t_r)\tilde{C}_{gd}
\end{align}
Therefore, \( \lambda''(t_r) = 0 \) if and only if Eq. \ref{appeq:theo3} holds.

Now, we can analyze the correlation between \( \lVert \delta \rVert \) and \( t_r^\ast \). Note that \( \tilde{C}_{gd} \) is the lower bound of \(2\lVert\nabla_x\log \rho_t(x)\rVert\). With the increase of disturbance \(\lVert \delta \rVert\), \( \tilde{C}_{gd} \) will subsequently increase, causing the right-hand side of Eq. \ref{appeq:theo3} to increase.

In our setting, \( \zeta(t_r) = \frac{1}{4}kt_r^2 \). Therefore, Eq. \ref{appeq:theo3} can be rewritten as:
\begin{align}
    -\frac{1}{2}kt_r^2(e^{\frac{1}{2}kt_r^2}-1)^{-3/2} + (e^{\frac{1}{2}kt_r^2}-1)^{-1/2} = -2\zeta(t_r)(e^{2\zeta(t_r)}-1)^{-3/2} + (e^{2\zeta(t_r)}-1)^{-1/2} = \frac{\tilde{C}_{gd}}{C_{\sigma,d}}
\end{align}
The left-hand side of Eq. \ref{appeq:theo3} is monotonically increasing with respect to \( \zeta(t_r) \) in the interval \([0, 1]\). Since \( \zeta(t_r) \) is also a monotonically increasing function with respect to \( t_r \), within a certain range, the left-hand side of the equation is a monotonically increasing function with respect to \( t_r \). In other words, an increase in \(\lVert \delta \rVert\) will lead to an increase in \( t_r^\ast \), indicating a positive correlation between \(\lVert \delta \rVert\) and \( t_r^\ast \).
\end{proof}

\begin{wrapfigure}{r}{0.38\textwidth}
    \centering
    \includegraphics[width=2.55in]{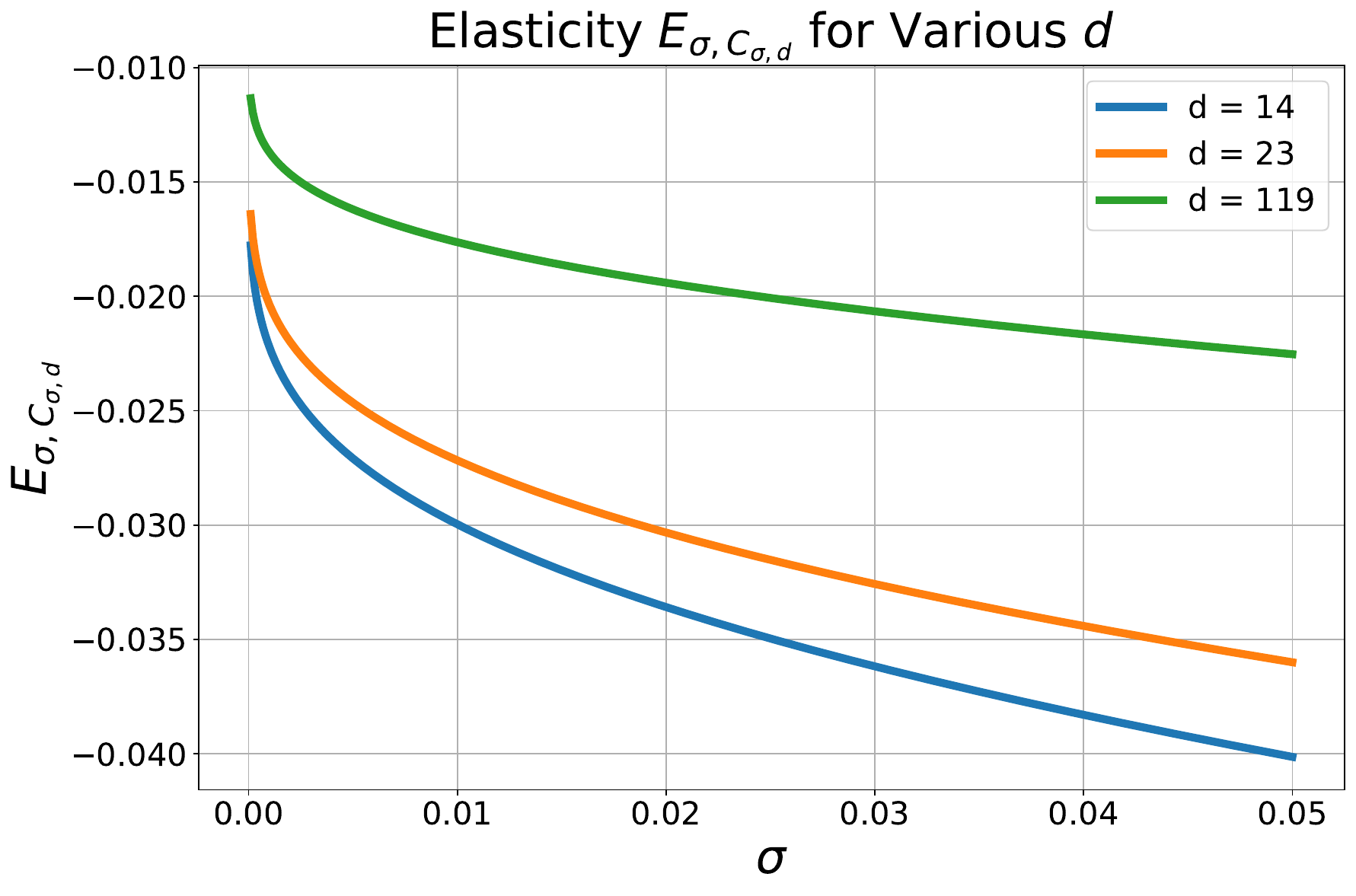}
    \vskip -0.1in
    \caption{The corresponding curve of the elasticity $E_{\sigma, C_{\sigma,d}}$ with respect to $\sigma$.}
    \label{appfig:sigma}
\end{wrapfigure}

\subsection{Sensitive Analysis of $\sigma$}
The lower bound of \( \lVert x_o-\hat{x_o} \rVert \) can be regarded as a function with four parameters: \( t_r, \sigma, d, \) and \( \delta \). However, once the optimal demonstration \( x_o \) and the corresponding sub-optimal demonstration \( x_s \) are provided, \( \delta \) and \( d \) are fixed. Furthermore, considering that different demonstrations exhibit varying tolerances for sub-optimal behavior, the magnitude of \( \sigma \) can be set according to actual scenarios. In most cases, \( \sigma \in [0.05, 0.0001] \). Therefore, $\lambda$ is indeed an increasing function with respect to \( t_r \) when a pair of demonstrations and a confidence level $1-\sigma$ are specified. However, it can be observed that among these four parameters, only $\sigma$ is subjectively determined by humans. We hope that for the same pair of optimal and sub-optimal demonstrations, the different subjective choices of each individual have little impact on the final calculated value of $t_r$. We provide a new theorem below to show that different confidence level $\sigma$ has little impact on the choice of $t_r$.
\begin{theorem}
    The selection of \( t_r^\ast \) exhibits low sensitivity to the magnitude of \( \sigma \), indicating that, for a purified demonstration, configuring different confidence levels has minimal impact on \( t_r^\ast \). Additionally, as the dimension of the demonstration $d$ increases, the sensitivity of $t_r^\ast$ to $\sigma$ will also decrease. 
    \label{app:theo5}
\end{theorem}
\begin{proof}
First of all, we will introduce elasticity, which is an indicator that can be used to measure the extent to which a change in one variable will affect other variables. Define elasticity $E_{x,y}$ of variables $x$ and $y$ as follows,
\begin{align}
    E_{x,y} = \frac{d(lny)}{d(lnx)}.
\end{align}
We have already known that $C_{\sigma,d} = \sqrt{2d(1+\sqrt{\frac{8ln(2/\sigma)}{d}})}$, then 
\begin{align}
    E_{\sigma, C_{\sigma,d}} = -\frac{1}{\sqrt{2dln(2/\sigma)} + 4ln(2/\sigma)}.
\end{align}
Since $1-\sigma$ represents the confidence level, in most cases, the value range of $\sigma$ is 0.05 to 0.0001. Note that the higher the accuracy requirement for the demonstration, the smaller the corresponding selected $\sigma$. In our experiment, the dimension of data is 14, 23 and 119, and the corresponding curve of the elasticity $E_{\sigma, C_{\sigma,d}}$ is in Figure \ref{appfig:sigma}.

From the figure, we can find that $\lVert E_{\sigma, C_{\sigma,d}} \rVert < 0.0113$ for every $d$, which reflects that $C_{\sigma,d}$ is not sensitive to the change of $\sigma$ , when the value range of $\sigma$ is 0.05 to 0.0001. In light of the analysis derived from Eq. 59, it is evident that with increasing complexity in the demonstration, as indicated by the augmentation of the data dimension $d$, the sensitivity of parameter $E_{\sigma, C_{\sigma,d}}$ to variations in parameter $\sigma$ exhibits a notable decrement. This observation implies that in scenarios involving more intricate demonstrations, particularly when the value of \( \sigma \) lies within the range of 0.05 to 0.0001, the selection of \( \sigma \) becomes substantially less consequential. This diminishing relevance of \( \sigma \) selection is a critical consideration in the context of increasingly complex data dimensions, as it underscores when dealing with a high-dimensional real-world data, the choice of different \( \sigma \) has a negligible impact on the optimal \( t_r^\ast \).
    
\end{proof}

\section{Experimental Details}
Our source code and training data will be available at \url{https://github.com/yunke-wang/dp-il}.

\subsection{Implementation Details of Diffusion Model}
The architecture of diffusion model $\epsilon_\phi$ is five linear layers with a 0.2 dropout ratio, batch normalization, and ReLU nonlinear activation, and the size of the hidden dimension is 1024. We implement the above two algorithms based on this repo (\url{https://github.com/abarankab/DDPM}). The training epoch is set to be 10000 and the learning rate of $\epsilon_\phi$ is set to 1e-4. We set $N=1000$ for all experiments and set the forward process variances to constants increasing linearly from $\beta_1=1e-4$ to $\beta_N=0.02$. The pseudo code of diffusion model's training and purification is available in Algorithm \ref{algo:diffusion_train} and Algorithm \ref{algo:diffusion_purification}.
\begin{minipage}[t]{.48\textwidth}
    \begin{algorithm}[H]
        \caption{Diffusion Model Training}
        \begin{algorithmic}[1]
            \REPEAT
                \STATE Sample optimal demonstrations $x_o \sim \mathcal{D}_o$
                \STATE $i\sim \mathrm{Uniform}(\{0,...,N-1\})$
                \STATE $\epsilon\sim\mathcal{N}(0,I)$
                \STATE Optimize diffusion model $\epsilon_\phi$ by taking gradient descent step on 
                \STATE $\nabla_\phi[\epsilon - \epsilon_\phi(\sqrt{\Bar{\alpha}_i}\cdot x_o + \sqrt{1-\Bar{\alpha}_i}\cdot\epsilon,i)]^2$
            \UNTIL converged
        \end{algorithmic}
        \label{algo:diffusion_train}
         \vskip 0.05in
    \end{algorithm}
\end{minipage}%
\hfill
\begin{minipage}[t]{.48\textwidth}
    \begin{algorithm}[H]
        \caption{Diffusion Purification}
        \begin{algorithmic}[1]
            \STATE Sample sub-optimal demonstrations $x_s \sim \mathcal{D}_s$;
            \STATE Calculate $x_{s,i^\ast}$ via forward diffusion: 
            \STATE $x_{s,i^\ast} = \sqrt{\Bar{\alpha}_i}\cdot x_s + \sqrt{1-\Bar{\alpha}_i}\cdot\epsilon$, $\quad \epsilon\sim\mathcal{N}(0, I).$
            \FOR{$i=i^\ast, ..., 1$}
                \STATE Calculate $x_{i-1}$ via reverse diffusion:
                \STATE $x_{s,i-1}=\frac{1}{\sqrt{\alpha_i}}\Big(x_{s,i} - \frac{1-\alpha_i}{\sqrt{1-\Bar{\alpha}_i}\epsilon_\phi(x_{s,i}, t)} \Big) + \sqrt{\beta_i} z,$
            \ENDFOR
            \STATE \textbf{Return} $\hat{x}_{s,0}$
        \end{algorithmic}
        \label{algo:diffusion_purification}
        \vskip -0.02in
    \end{algorithm}
\end{minipage}

\subsection{Implementation Details of Imitation Learning}
We use a deep neural network that has two $100\times100$ fully connected layers and uses Tanh as the activation layer to parameterize policy. To output continuous action, agent policy adopts a Gaussian strategy, hence the policy network outputs the mean and standard deviation of action. The continuous action is sampled from the normal distribution formulated with the action's mean and standard deviation.
For online imitation learning methods, the discriminator and value function are using the same architecture as the policy network. 

In offline imitation learning, the policy is trained with batch size 256, and the total epoch is set to be 1000. For online imitation learning, the learning rate of the discriminator $D_\psi$ and the critic $r_\psi$ is set to $3 \times 10^{-4}$. Five updates on the discriminator follow with one update on the policy network in one iteration. For the value function, the learning rate is set to $3 \times 10^{-4}$ and three training updates are used in one iteration. We conduct the on-policy method TRPO \cite{schulman2015trust} as RL step in online imitation learning, the learning rate is set to $3\times 10^{-4}$ with batch size 5000. The discount rate $\gamma$ of the sampled trajectory is set to 0.995. The $\tau$ (GAE parameter) is set to 0.97. 

\subsection{Data Collection}
We provide six supplementary sub-optimal demonstration datasets to evaluate the performance of DP-IL on 4 MuJoCo tasks. 
After we train an optimal policy $\pi_o$ by TRPO, we use \textbf{two} different kinds of ways to collect sub-optimal demonstration data in our experiments. The quality of sub-optimal demonstrations is provided in Table \ref{data_quality}. 

The first way to collect sub-optimal demonstrations is to add different Gaussian noise $\xi$ to the action distribution $a^\ast$ of $\pi_s$ to form sub-optimal policy $\pi_s$. The action of $\pi_s$ is modeled as $a\sim\mathcal{N}(a^\ast, \xi^2)$ and we choose $\xi=[0.6, 0.4, 0.25]$ to generate sub-optimal demonstrations datasets (\textit{e.g.}, D1-L1, D1-L2 and D1-L3).  This way of collecting sub-optimal demonstration has been used in several works~\cite{sasaki2021behavioral, tangkaratt2019vild}.

The second way to collect sub-optimal demonstrations is to save checkpoints during the RL training of $\pi_s(a|s)$. In our experiment, the RL training is conducted with 5M interactions with the environment. We save 3 checkpoints at 1M, 1.5M and 3M interactions to sample D2-L1, D2-L2 and D2-L3 datasets. While the agent converges fast
This way of collecting sub-optimal demonstration has been investigated in ~\cite{wang2021learning, wu2019imitation}.

\begin{table}[!h]
	\centering
	\caption{The quality of demonstrations in 4 MuJoCo tasks, which is measured by the average cumulative reward of trajectories.}
    \vskip 0.05in
	\begin{tabular}{lll|ccc|ccc|c}
		\toprule
		Task& $\mathcal{S}$ &$\mathcal{A}$ & D1-L1 & D1-L2 & D1-L3 & D2-L1 & D2-L2 & D2-L3  &\bf Expert\\
		\toprule
		Ant-v2         & $\mathbb{R}^{111}$ & $\mathbb{R}^{8}$  & -73 & 227 & 3514 & 1062 & 1560 & 2649  &4349  \\
		HalfCheetah-v2 &$\mathbb{R}^{17}$&$\mathbb{R}^{6}$ & 567 & 1090 & 1853 & 1491 & 2217 & 3263  &4624\\
  		Walker2d-v2    &$\mathbb{R}^{17}$&$\mathbb{R}^{6}$ & 523 & 467 & 4362 & 1699 & 2717 & 4152   &4963 \\
		Hopper-v2      &$\mathbb{R}^{11}$ &$\mathbb{R}^{3}$ & 699 & 1037 & 3229 & 734 & 3362 &  2597 &3594 \\
		\bottomrule
	\end{tabular}
 \label{data_quality}
\end{table}

\subsection{Compared Methods}
We compare DP-IL with several \textit{state-of-the-art} imitation learning with imperfect demonstration methods.
Specifically, several offline imitation learning methods (\textit{i.e.}, BCND~\cite{sasaki2021behavioral}, DWBC~\cite{xu2022discriminator} and DemoDICE~\cite{kim2021demodice}) and online imitation learning methods (\textit{i.e.}, 2IWIL \cite{wu2019imitation}, IC-GAIL \cite{wu2019imitation} and WGAIL \cite{wang2021learning}) are compared.
We briefly review the details of methods compared against DP-IL in our experiments below.

\noindent\textbf{BCND, DWBC and DemoDICE}\quad
We follow the instruction in BCND's paper to implement BCND. Notice that for a fair comparison with other offline imitation learning methods, we do not use ensemble policies for BCND. As for DWBC, we adapt their released code\footnote{\url{https://github.com/ryanxhr/DWBC}} to conduct experiments. As for DemoDICE, we adapt the implementation in SmoDICE repo\footnote{\url{https://github.com/JasonMa2016/SMODICE/tree/main}} to conduct experiments.

\noindent\textbf{2IWIL and IC-GAIL}\quad
We re-implement 2IWIL/IC-GAIL based on the official implementation\footnote{\url{https://github.com/kristery/Imitation-Learning-from-Imperfect-Demonstration}}. 
In 2IWIL and IC-GAIL, a fraction of imperfect expert demonstrations are labeled with confidence (\textit{i.e.}, $\mathcal{D}_l=\{(s_i, a_i), r_i\}^{n_l}_i$), while the remaining demonstrations are unlabeled (\textit{i.e.}, $\mathcal{D}_u=\{(s_i,a_i)\}^{n_u}_i$). Since we have no access to the confidence score of the state-action pair in our setting, we use the normalized reward of the demonstrator as the confidence score for their related demonstrations. In the experiment, we choose 20\% labeled demonstrations to train a semi-supervised classifier and then predict confidence for other 80\% unlabeled demonstrations.

\noindent\textbf{WGAIL}\quad
WGAIL is proposed to estimate confidence in GAIL framework without auxiliary information. The confidence $w(s,a)$ of each demonstration is calculated by $[(1/D^\ast_w(s,a)-1)\pi_\theta(a|s)]^{\frac{1}{\beta+1}}$. The confidence estimation and GAIL training interact during the training. Followed with the official implementation\footnote{\url{https://github.com/yunke-wang/WGAIL}}, $\beta$ is set to be 1.

\subsection{Additional Experimental Results}
\subsubsection{Impact of reverse point $i_r$ in RoboSuite}
We also include Robosuite results to offer a more complete understanding of how $i_r^\ast$ impacts demonstrations of varying quality. In Robosuite, we utilized 10 trajectories as expert demonstrations, as shown in Table \ref{apptab:trajid}. Among these, two high-reward trajectories (ID 6 and 10) were considered optimal demonstrations, while the rest were designated as sub-optimal demonstrations. To assess the impact of different demonstration qualities, we divided the sub-optimal demonstrations into two groups, Robo-L1 and Robo-L2.
\begin{table}[!h]
    \centering
        \vskip -0.05in
    \caption{The length and quality of 10 trajectories in RoboSuite.}
    \vskip 0.05in
    \begin{tabular}{c|cccccccccc}
    \toprule
       Traj\_ID  & 1 & 2 & 3& 4& 5& 6& 7& 8& 9& 10  \\
       \midrule
       Length  & 500&502&502&505&506&507&507&509&511&511\\
       Reward &276&120&203&102&253&312&103&81&180&318 \\
       \bottomrule
    \end{tabular}
    \label{apptab:trajid}
\end{table}

For Robo\_L1, we selected four trajectories with rewards below 150 (ID 2, 4, 7, and 8) as sub-optimal demonstrations. For Robo-L2, we chose four trajectories with rewards ranging from 150 to 300 (ID 1, 3, 5, and 9) as sub-optimal demonstrations. Clearly, Robo-L2 demonstrates higher quality compared to Robo-D1. We assessed the influence of timestep $i_r^\ast$ on both Robo-L1 and Robo-L2. The results, as shown in Table \ref{apptab:robosuite}, exhibit similar trends to those observed in the MuJoCo tasks.
\begin{table}[!h]
    \centering
    \vskip -0.05in
    \caption{Impact of $i_r$ in RoboSuite platform with human demonstrations.}
    \vskip 0.05in
    \setlength{\tabcolsep}{7pt} 
    \begin{tabular}{c|ccccc}
    \toprule
     $i_r$   & 1& 3 & 5 & 10 & 30 \\
    \midrule
    DP-BC (Robo-L1) & 0.60 & 0.74 &\textbf{0.80} & 0.64 & 0.48  \\
    DP-BC (Robo-L2) & \textbf{0.82} & 0.78 & 0.76 & 0.76 & 0.52 \\
    \bottomrule
    \end{tabular}
    \label{apptab:robosuite}
\end{table}

\subsection{Impact of reverse point $i_r$ in MuJoCo}
We have provided the ablation study on $i_r$ with D1 demonstrations in Figure \ref{fig:optimal_t}, and we provide full results of $i_r$ with both D1 and D2 demonstrations in Table \ref{tab:optimal_t}. 
From the table, we can observe that in most cases (19 out of 24 cases), setting $i_r$ within the 5-10 range consistently leads to optimal or near-optimal performance. These cases are across various tasks and demonstration qualities, which indicates that an $i_r$ value in this range can serve as a robust default choice that yields satisfactory results in most situations.
\begin{table}[!h]
    \centering
    \caption{Performance of DP-BC when setting different $i_r$, which is measured by the average cumulative reward of 10 trajectories. The results are related to Figure \ref{fig:optimal_t}.}
    \vskip 0.05in
    \begin{tabular}{c|c|ccc|ccc}\toprule
     Task   &$i_r$ & D1-L1 & D1-L2 & D1-L3  & D2-L1 & D2-L2 & D2-L3   \\ \midrule
    \multirow{5}{*}{Ant-v2}  &100 & \bf 261±54 & 380±62 & 486±76 & 199±66 & 342±14  &  328±88 \\
        &50 & 202±51 & \bf 803±114& 1073±172 & 339±27 & 720±139 &1776±123 \\
        &30 & 47±16 & 719±51& 1890±156 & \bf 1572±160 & 1833±30&3064±34 \\
        &10 & -100±51 & 499±77& \bf 2547±118 & 1402±151 & 1982±111 &3260±34 \\
        &5 & -133±55 & 490±125& 2331±248 & 1105±263 & \bf 1984±103 & \bf 3414±40 \\ \bottomrule
    \multirow{5}{*}{HalfCheetah-v2}  &100 & 556±102 & 640±108 & 1115±181 & 1123±41 & 1791±13 &1393±175 \\
        &50 & 797±185  & 1896±222 & 2733±113 & 1468±82 & 2421±5 &3126±306 \\
        &30 & 895±73   & 1880±326 & 2899±257 & \bf 1530±39 & 2714±15 &4434±240 \\
        &10 & \bf 1365±147 & \bf 2440±274 & \bf 4042±80& 1394±103 & \bf 2749±15 &\bf 4748±40 \\
        &5 & 708±255 & 2061±373 & 3758±351 & 1469±253 & 2469±253 & 4660±14 \\ \bottomrule
    \multirow{5}{*}{Walker2d-v2}  &100 & 260±2 & 241±12 & 311±9 & 259±1 & 258±1 &273±19  \\
        &50 & 463±40 & 259±11& 553±89 & 549±196 & 360±14 &1085±251  \\
        &30 & 415±56& 978±96& 1398±333 & \bf 2208±160 & 1742±253 &2512±212  \\
        &10 & \bf 1697±219& 1267±145& 1474±265 & 1764±143 & 1915±384 &1984±365  \\
        &5 & 1467±253 & \bf 1722±297& \bf 3020±466 & 1917±151 & \bf 3162±20 &\bf 3076±205  \\ \bottomrule
    \multirow{5}{*}{Hopper-v2}  &100 & 98±2 & 461±11 & 420±1 & 215±1 & 151±3 & 355±25 \\
        &50 & 516±0 & 815±16 & 379±0 & 1283±109 & 631±4 & 342±29 \\
        &30 & 688±33 & 649±43 & 517±1 & 2075±159 & 904±6 & 1353±128\\
        &10 & \bf 1000±42 & 2752±38 & 678±56 & \bf 2308±113 & \bf 2323±2 & 1740±175\\
        &5 & 580±60 & \bf 3145±11 & \bf 1508±272 & 1835±216 & 2297±34 & \bf 2265±171 \\ \bottomrule
    \end{tabular}
    \label{tab:optimal_t}
\end{table}

While in most cases, an $i_r$ setting within the range of 5-10 yields optimal performance, we can also observe the selection of $i_r$ has a connection to the task. In the 4 MuJoCo tasks used in the experiment, Ant-v2 is the most challenging task, followed by HalfCheetah-v2 and Walker2d-v2, with Hopper-v2 being the easiest. We can observe that the average optimal $i_r$ is relatively higher (for instance, around 100 for D1-L1) for Ant-v2 and smaller for Hopper-v2, where an optimal $i_r$ is as low as 5 for all datasets. This suggests that setting a relatively higher $i_r$ for more difficult tasks is reasonable. Following the results presented in Table \ref{tab:optimal_t}, an empirical way is to set $i_r$ from 5 to 10 for simple tasks like Hopper-v2, and set $i_r$ from 10 to 50 for challenging tasks like Ant-v2.




\end{document}